\documentclass[journal,twoside]{IEEEtran}
\usepackage[noadjust]{cite}

\usepackage{color,soul}
\soulregister\cite7 
\soulregister\citep7 
\soulregister\citet7 
\soulregister\ref7 
\soulregister\pageref7 

\usepackage{soul}  
\usepackage{xcolor}  

\ifCLASSINFOpdf
  \usepackage[pdftex]{graphicx}
  \graphicspath{{./pdf/}{./jpeg/}}
  \DeclareGraphicsExtensions{.pdf,.jpeg,.PNG}
\else
  \usepackage[dvips]{graphicx}
  \graphicspath{{./eps/}}
  \DeclareGraphicsExtensions{.eps}
\fi
\usepackage{xcolor}
\usepackage[switch]{lineno}
\usepackage{amsmath,amssymb,amsthm,mathtools}
\usepackage{algorithm}
\usepackage{algorithmic}
\usepackage{bm}
\usepackage{multirow}
\usepackage{soul}
\usepackage{makecell}

\begin{document}
%
\title{
Programmable Locking Cells (PLC) for Modular Robots with High Stiffness Tunability and Morphological Adaptability
} 
%
%

\author{Jianshu 
       Zhou$^{1}$\, \IEEEmembership{Member,~IEEE},
        Wei Chen$^{2}$, 
        Junda Huang$^{2}$,
        Boyuan Liang$^{1}$, \IEEEmembership{Student Member,~IEEE},
        \\Yunhui Liu$^{2}$,~\IEEEmembership{Fellow,~IEEE},
        and Masayoshi Tomizuka$^{1}$,~\IEEEmembership{Life Fellow,~IEEE} 
        
\thanks{ Research supported in part of the HK RGC under GRF-14207423, in part by the InnoHK of the Government of Hong Kong via the Hong Kong Centre fozr Logistics Robotics, and in part by the VC Fund 4930745 of the CUHK T Stone Robotics Institute.} 
\thanks{$^{1}$Authors with the Department of Mechanical Engineering, University of California, Berkeley.}
\thanks{$^{2}$Authors with the Department of Mechanical and Automation Engineering, The Chinese University of Hong Kong.}
}

\maketitle


\begin{abstract}

Robotic systems operating in unstructured environments require the ability to switch between compliant and rigid states to perform diverse tasks such as adaptive grasping, high-force manipulation, shape holding, and navigation in constrained spaces, among others. However, many existing variable stiffness solutions rely on complex actuation schemes, continuous input power, or monolithic designs, limiting their modularity and scalability. This paper presents the Programmable Locking Cell (PLC)—a modular, tendon-driven unit that achieves discrete stiffness modulation through mechanically interlocked joints actuated by cable tension. Each unit transitions between compliant and firm states via structural engagement, and the assembled system exhibits high stiffness variation—up to 950\% per unit—without susceptibility to damage under high payload in the firm state. Multiple PLC units can be assembled into reconfigurable robotic structures with spatially programmable stiffness. We validate the design through two functional prototypes: (1) a variable-stiffness gripper capable of adaptive grasping, firm holding, and in-hand manipulation; and (2) a pipe-traversing robot composed of serial PLC units that achieves shape adaptability and stiffness control in confined environments. These results demonstrate the PLC as a scalable, structure-centric mechanism for programmable stiffness and motion, enabling robotic systems with reconfigurable morphology and task-adaptive interaction.

\end{abstract}


%

\section{Introduction}
%
%
%
%

\IEEEPARstart{R}{obotic} systems deployed in unstructured environments need to dynamically balance compliance and rigidity to meet diverse task demands. Tasks such as navigating constrained spaces, manipulating fragile objects, or performing high-force operations require robots to adapt both their morphology and structural stiffness. This tunable mechanical behavior is particularly critical in applications ranging from surgical instruments~\cite{kim2013mri}, adaptive grippers and manipulators~\cite{shan2023variable, manti2016review}, and wearable or assistive devices~\cite{polygerinos2017soft} to field-deployable mobile platforms~\cite{jiang2012granular}.

Existing variable stiffness solutions enable stiffness change but often face challenges in real-world robotic applications. These include actuation complexity, limited modularity, and narrow stiffness tuning ranges~\cite{wolf2015review, manti2016review, dou2021soft}. Moreover, many systems only provide coarse or binary stiffness modulation—typically switching between “soft” and “locked” states—without fine adjustability. In such designs, the locked configurations are often vulnerable to damage or failure under high payloads due to their reliance on active force balancing or non-structural stiffening mechanisms. These limitations hinder robots from simultaneously achieving high morphological flexibility, robust stiffness modulation, and safe operation under varying load conditions, thereby constraining their ability to perform complex configurations and adapt to diverse task requirements.

Modular robotic systems have demonstrated notable capabilities in self-reconfiguration, distributed control, and scalable morphology~\cite{yim2000polybot, yim2007modular, seo2019modular}. Such systems offer structure-level adaptability well suited for dynamic environments. However, most modular platforms focus on topological transformation and largely overlook localized stiffness programmability—an essential capability for robots to interact stably with physical environments and execute high-force tasks.

To address these challenges, we propose a structure-centric solution: the Programmable Locking Cell (PLC), a modular tendon-driven unit that enables discrete stiffness modulation via mechanical self-locking. Each PLC unit transitions between compliant and firmed states through tendon-actuated engagement of interlocking teeth, and a single unit achieves up to 950\% stiffness variation. We develop a comprehensive analytical framework capturing both the firmed stiffness, modeled via elastic beam theory, and loosening stiffness, characterized by tendon deformation under overload-induced detachment. Directional stiffness anisotropy, overload transitions, and torsional resistance are analyzed and validated through single- and multi-segment experiments. We further introduce a discrete inverse kinematics method based on a k-d tree search to address point reaching within the robot’s non-continuous workspace. Together, these models and analyses establish the physical basis and control feasibility of the PLC system.

To demonstrate its practical versatility, we present two functional robotic prototypes. The first is a two-fingered variable-stiffness gripper, where each finger consists of three serially connected PLC units that support both independent and coordinated rotation. This configuration enables adaptive grasping, firm holding, and in-hand manipulation. The second prototype is a 16-segment pipe-traversing robot, in which the rear 13 segments share unified stiffness control to passively conform to narrow and complex geometries, while the front 3 segments employ independent stiffness tuning to perform precise configuration adjustments and execute distal tasks. These demonstrations highlight the PLC as a scalable and structure-centric mechanism for programmable stiffness and morphology. In particular, the modular tendon-driven design provides intrinsic overload tolerance in the firmed state, enhancing structural robustness and enabling safe high-force interaction in unstructured environments.

The contributions of this work are summarized in four parts:
\begin{itemize}
  \item We propose a modular tendon-driven unit for programmable stiffness and motion. The Programmable Locking Cell (PLC) achieves discrete stiffness transitions through tendon-actuated mechanical locking, enabling up to 950\% stiffness variation with independent or coordinated motion control.

  \item We present a scalable architecture for stiffness and morphological programmability. PLC units can be serially assembled into continuum-like structures with spatially distributed stiffness and reconfigurable shapes, as demonstrated in a 16-segment pipe-traversing robot and a two-fingered gripper composed of three PLC units per finger. These prototypes enable task-adaptive configurations for compliant navigation and precise object manipulation, respectively.

  \item We develop a unified modeling framework and validate it experimentally. Analytical models for both firmed and loosening stiffness states are derived using beam mechanics and tendon deformation, and verified through single- and multi-segment experiments.

  \item We introduce a k-d tree-based discrete path search method, which enables efficient motion planning in PLC's workspaces.
\end{itemize}

The remainder of the paper is organized as follows. Section II reviews related work and situates the proposed PLC mechanism within the landscape of variable stiffness robotic systems. Section III details the structural design, actuation principles, and programmable stiffness mechanism of the PLC unit and its multi-unit assembly. Section IV presents the kinematic modeling and search-based motion planning method for discrete configuration spaces. Section V introduces a stiffness normalization metric, models the firmed and loosening stiffness of the PLC robot, and experimentally characterizes its stiffness anisotropy and trajectory tracking performance. Section VI provides experimental validation of the stiffness models, motion control, and functional demonstrations, including adaptive grasping and constrained-space manipulation. Section VII concludes the paper and outlines directions for future work.

\section{Related Works}

\begin{figure}[!t] 
    \centering
    \includegraphics[width=\linewidth]{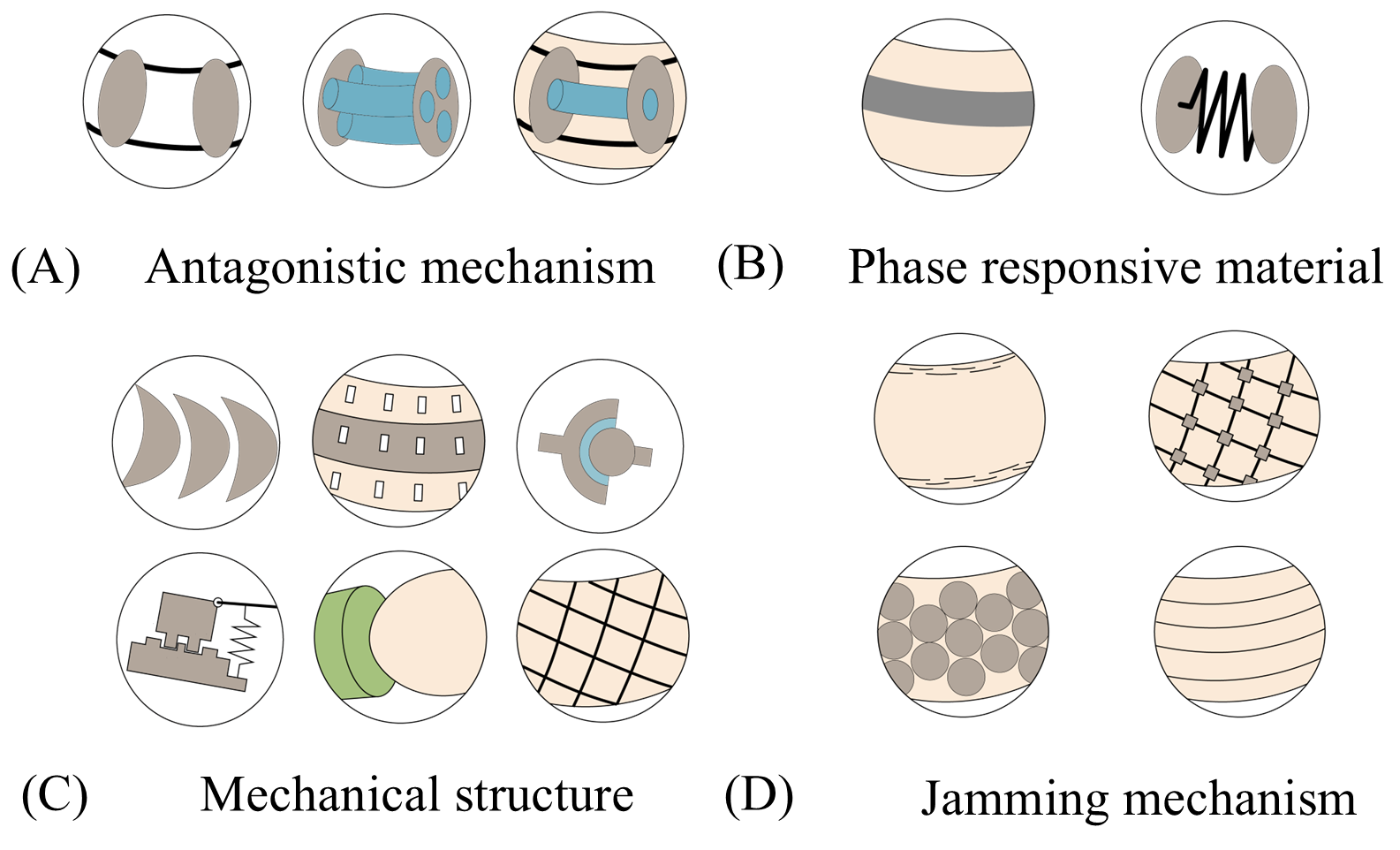}
    \caption{
        Different variable stiffness design schemes in continuum robots. (A) Typical antagonistic mechanism, including tendon method, fluidic method, and mixed method. \cite{rao2021model, toshimitsu2021sopra, shiva2016tendon} (B) Phase responsive material to change stiffness: fluidic metal and shape memory alloy. \cite{kim2017active, an2023active} (C) Variable stiffness using mechanical structures, including semi-sphere structure, concentric anisotropic tube, ball-joint based structure, lockable mechanism, tube-constrain-based structure, net and soft skin constrain structure. \cite{lin2022modular, kim2019continuously, li2015ultimate, jitosho2023passive, kim2017novel} (D) Jamming mechanism, including granular, layer, surface, and fabrics structure jamming \cite{kim2013novel, hassan2017active, fan2022novel}.
    }
    \label{fig:continuumrobotmechanism}
\end{figure}

Achieving programmable stiffness modulation across a broad range is vital for robotic systems operating in unstructured environments. Existing approaches to variable stiffness design can be broadly categorized by their underlying principles: antagonistic actuation, phase-change materials, jamming-based systems, and mechanically or structure-centric designs. Each method presents unique trade-offs in terms of control complexity, energy efficiency, stiffness range, and structural scalability. This section reviews these strategies and identifies their limitations in relation to our objective: achieving both high stiffness variation and programmable morphology using modular, structure-centric design.

\textbf{Antagonistic actuation} is inspired by the biological principle of muscles generating stiffness through opposing forces. This approach includes tendon-based~\cite{kim2013stiffness}, fluid-based~\cite{al2017design}, and tendon-fluid hybrid configurations~\cite{shiva2016tendon}, offering continuously tunable stiffness by adjusting internal tensions. While widely used in soft robotic arms and continuum manipulators~\cite{stilli2014shrinkable}, these methods often require complex feedback control and exhibit instability under high force loads. Moreover, the achievable stiffness range is often limited due to constraints on cable routing and structural buckling.

\textbf{Phase-change material strategies} employ smart materials such as SMAs~\cite{kim2017active}, SMPs~\cite{choi2020design,yang20163d}, and LMPAs~\cite{wei2021analysis} to switch between soft and stiff states. These methods are thermally activated and highly compact, making them attractive for minimally invasive surgical tools and deployable structures~\cite{saavedra2013variable}. However, their response time is slow due to thermal diffusion, and many phase-change materials offer only binary states with limited intermediate tuning. Additionally, actuating hardware (e.g., heaters or cooling elements) may increase system complexity and energy requirements.

\textbf{Jamming-based mechanisms} modulate stiffness by controlling internal friction in granular~\cite{zhou2022bioinspired, zhou2020adaptive}, fiber~\cite{arleo2023variable}, layered~\cite{kim2013novel, zeng2020parallel}, or hybrid media~\cite{clark2019assessing}. These designs are low-cost, lightweight, and capable of fast switching. They are widely implemented in soft grippers and reconfigurable interfaces. However, jamming systems often suffer from limited repeatability, insufficient stiffness range, and difficulty restoring to initial states~\cite{aktacs2021modeling}. Their performance also depends heavily on packing density and vacuum system stability, making them less reliable in dynamic tasks.

\textbf{Structure-centric and mechanically tunable mechanisms} vary stiffness through interlocking, constraint-based coupling, or friction modulation. Representative examples include concentric anisotropic tubes~\cite{kim2019continuously}, constraint-based continuum links~\cite{li2019flexible}, clearance-tunable ball joints~\cite{shen2023design}, and modular locking segments~\cite{lin2022modular, yang2020geometric}. These systems offer discrete stiffness states with strong support capacity and minimal energy consumption once locked. Due to their modularity and mechanical reliability, they are increasingly adopted in reconfigurable arms, pipeline inspection robots, and deployable tools. However, most designs only enable binary or coarse stiffness transitions and lack spatial stiffness programmability along the robot body.

A growing body of work has begun exploring modular tendon-driven units for achieving distributed stiffness control~\cite{lin2022modular}. These units often incorporate passive locking mechanisms to maintain stiffness without continuous actuation. Yang et al.~\cite{yang2020geometric} proposed shape memory alloy-initiated locking for morphologically constrained structures, enabling shape holding with high payloads. Nonetheless, many such approaches remain constrained to fixed geometries, lack programmable morphing behavior, or offer limited stiffness variation ranges.

\begin{figure}
\centering
    \includegraphics[width=\linewidth]{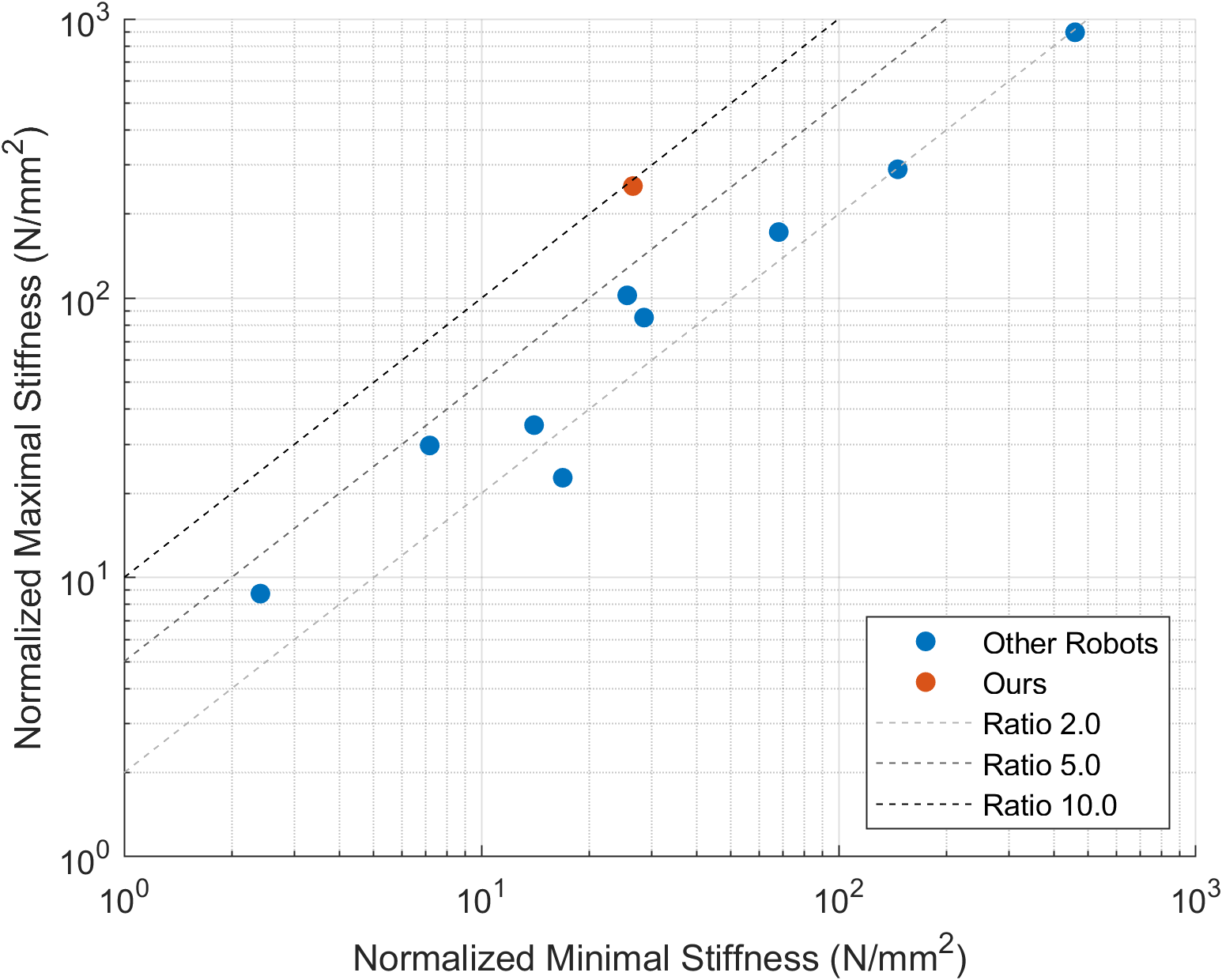}
\caption{Summary of the normalized minimal and maximal stiffness in recent approaches to varying stiffness robots. The raw data, references and normalization method can all be found in Appendix \ref{apdx:notmalized stiffness}. A design has a larger stiffness range if its corresponding point is closer to the upper left corner. Notice that both x and y-axis are log-scaled, indicating that a slight move to the left or up means a multiple fold increase in the minimal stiffness or the maximal stiffness. The three dashed lines stand for regions when the maximal stiffness is two, five or ten times the magnitude of the maximal stiffness. We can see that our approach is the most upper-left and the closest to the 10.0 ratio reference line.}
\label{fig:summarize scatter}
\end{figure}

\begin{table}[t]
\begin{center}
\caption{CUSTOMIZED PARAMETERS OF THE PLC ROBOT}
\label{table: major parameters}
\begin{tabular}{|p{4cm}|p{3cm}|}
    \hline   
    \multicolumn{1}{|c}{\centering $Parameters$} &
    \multicolumn{1}{|c|}{\centering $Value$} \\
    \hline   
    Young's modulus $E$ & 100$\sim $130 (MPa)\\
    \hline
    Spine outer diameter & 8 ($mm$) \\
    \hline
    Spine inner diameter & 2 ($mm$) \\
    \hline
    Spline length $l$ & 30 ($mm$) \\
    \hline
    Inclined angle $\beta$ & 30 ($^{\circ}$) \\
    \hline
    Origami skin outer diameter $D$ & 22 ($mm$) \\
    \hline
     Origami skin inner diameter $d$ & 17 ($mm$) \\
    \hline
\end{tabular}
\end{center}
\vspace{-0.2cm}  
\end{table}

Compared to these prior works, our approach introduces a structurally driven stiffness modulation mechanism based on a \emph{Programmable Locking Cell} (PLC). Each unit uses tendon-actuated locking rings with mechanical teeth to discretely switch between flexible and rigid states. The design supports a quasi-passive hold with no power consumption in the locked state, enabling scalable use in long-hold applications. Moreover, the PLC can be assembled serially, allowing task-specific morphology and spatial stiffness programming. As shown in Fig.~\ref{fig:summarize scatter}, our system achieves the highest stiffness variation ratio (9.5) among surveyed systems, according to geometric-normalized metrics described in Appendix~\ref{apdx:notmalized stiffness}.

\section{Design}
\label{sec:design}

This section provides a comprehensive overview of the principles and design of the PLC unit, covering its structural composition, motion mechanisms, variable stiffness functionality, and programmable motion capabilities. The motion mechanisms include the principles of rotation, actuation, and degrees of freedom, which together explain how and why the PLC units move. The variable stiffness mechanism illustrates both the implementation and results of stiffness modulation, while the programmable motion section describes how a PLC robot composed of multiple units can achieve complex, task-specific movements and behaviors.

By connecting multiple PLC units in series, different robotic components can be realized, such as a continuum manipulator or a robotic finger. To intuitively present the motion principles and variable stiffness mechanism of the PLC unit, we first explain the design and principles using a straight PLC unit, followed by an explanation of the PLC robot composed of inclined PLC units. The key difference between an inclined unit and a straight unit is the presence of an inclination angle, which enables the PLC robot to achieve spatial positioning. The key design parameters of the PLC unit are listed in Table~\ref{table: major parameters}.

\begin{figure}[!t] 
    \centering
    \includegraphics[width=\linewidth]{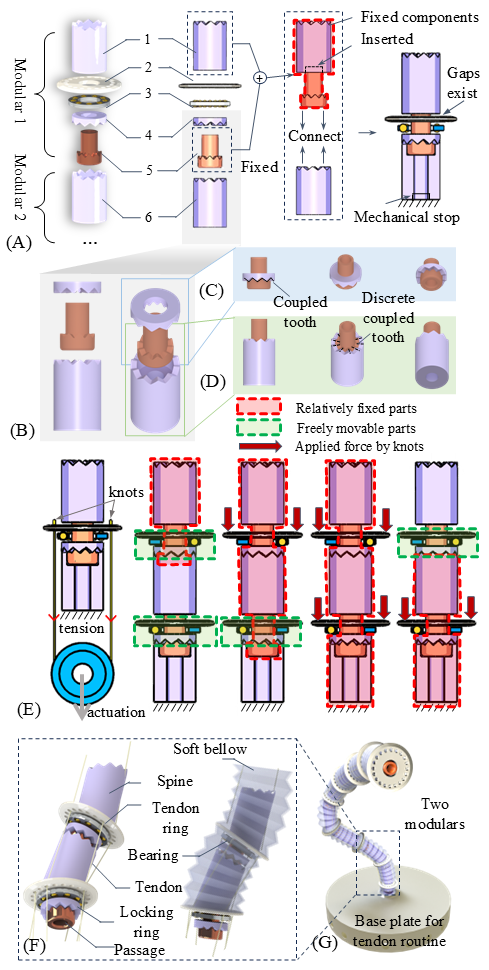}
    \caption{
        The structural composition of PLC units and variable stiffness mechanism of multiple linear PLC units. (A) The structure of PLC unit. \#1 to \#6 represent upper spine, tendong ring, bearing, lower spine, and upper spine respectively. During assembly, the lower spine is inserted into the upper spine and stopped by the mechanical stage. (B) The spine connection of two units. (C) Couple mechanism between the element 4 and 5 of the unit 1. (D) Couple mechanism between the element 5 and 6. Unit 1 and 2 are not rigidly fixed. (E) The tendon driven mechanism for stiffness variation. (F) Two connected PLC units. (G) A multi-segments PLC robot.
    }
    \label{fig:components_locking}
\end{figure}

\subsection{Structural Composition}

The PLC unit (Fig.~\ref{fig:components_locking}A) consists of five independent structural components: upper spine~(\#1), tendon ring~(\#2), bearing~(\#3), locking ring~(\#4), and lower spine~(\#5). The assembly of the PLC robot involves three steps: (1) assembling components \#1 to \#5 into a PLC unit, (2) connecting multiple PLC units in series using tendons to form a PLC robot, and (3) installing the origami skin onto each unit.

In the first step, the column of the lower spine passes through components \#2 to \#4 (annular structures) and is finally inserted into the upper spine (bottom view of the upper spine shown in Fig.~\ref{fig:components_locking}D). A mechanical stage is inside the upper spine, such that the lower spine can be mechanically stopped at the specific place. The contact between the upper spine and lower spine is bonded with strong adhesive, forming a fixed component (Fig.~\ref{fig:components_locking}A). This method constrains components \#2 to \#4 within the spine but allows them to move freely along the column of the spine in both vertical and rotational directions due to small gaps.

In the second step, the lower spine of the first unit is inserted into the upper spine of the second unit (Fig.~\ref{fig:components_locking}B). The upper spine of the second unit corresponds to component \#6 in Fig.~\ref{fig:components_locking}A. Unlike step one, where adhesive is used for fixation, the connection between adjacent units relies solely on stacking. The spine interface adopts a tight-tolerance slip-fit with a radial clearance of approximately 0.2 mm, produced by SLA printing and followed by light surface finishing at the engagement surfaces. This clearance provides sufficient compliance in the loosening state while maintaining full-tooth engagement without noticeable backlash in the locked state. Clearances significantly larger than this value were observed to introduce perceptible backlash even under tendon tension. After stacking the units, tendons pass through the tendon rings of both units. By following steps one and two iteratively, a multi-unit PLC robot can be assembled (Fig.~\ref{fig:components_locking}E).

In the third step, an origami skin is installed between the tendon rings of the first and second units (Fig.~\ref{fig:components_locking}F). The origami skin is rigidly attached to both the previous and subsequent tendon rings, ensuring that there is no relative rotation between the tendon rings. The origami skin serves three main purposes: (1) resisting torsional deformation caused by external and interal forces, (2) enabling bending towards different directions of inclined units, and (3) preventing buckling. The specific reasons for these functions are explained in the following two subsections.

By assembling inclined PLC units using the aforementioned three steps, a PLC robot can be obtained, as shown in Fig.~\ref{fig:components_locking}G. The main structure of the PLC robot consists of inclined units, with a base plate at the bottom for mounting pulleys and routing tendons. The passage at the center of the spine is a through-hole that allows flexible shafts or cables to pass through the PLC robot and function at the distal end.

\subsection{Motion Mechanism}
\label{subsec:motion mechanism}
To systematically explain how the PLC robot is controlled, this subsection is divided into three hierarchical topics: rotation principles, actuation principles, and degrees of freedom principles. The rotation principles describe how relative rotation occurs between adjacent units and the constraints governing this motion. The actuation principles explain how each degree of freedom is actuated. The degrees of freedom principles define the degrees of freedom available in both the PLC unit and the PLC robot, as well as how these degrees of freedom enable different behaviors.

Similar to the design section, the motion principles are first explained using straight PLC units for clarity, followed by an explanation of inclined PLC units. It is important to note that the motion principles and force interactions of straight and inclined units are identical.

\subsubsection{Rotation Principles}

The rotation between PLC units involves three components (Fig.~\ref{fig:components_locking}B): the lower spine of Unit 1, the locking ring, and the upper spine of Unit 2. These three components all have radial teeth that can interlock with each other. To describe the rotation mechanism of the PLC unit, we define two terms: coupled and decoupled. The coupled state indicates that the rotation between two units is synchronized; that is, if Unit 1 rotates by $x^\circ$, Unit 2 also rotates by $x^\circ$. Conversely, the decoupled state means that the rotation of one unit does not affect the other. The transition between coupled and decoupled states is determined by two meshing configurations of the three components mentioned above (Figs.~\ref{fig:components_locking}C and D).

Since the teeth of the locking ring face downward and the teeth of the lower spine face upward, they can interlock (Fig.~\ref{fig:components_locking}C). Although the locking ring can rotate continuously within 360 degrees, the number of teeth $N$ dictates its stable positions, which are discrete and can only occur at 360/$N \times n$, where $n \in \mathbb{Z}, 0 < n \leq N$. On the other hand, the teeth of the upper spine of Unit 2 face upward, aligning with the teeth of the lower spine of Unit 1 (Fig.~\ref{fig:components_locking}D). Defining the tooth surface areas of the lower spine, locking ring, and upper spine as $S_l$, $S_{lr}$, and $S_u$ respectively, we have the relation $S_l + S_u = S_{lr}$.

When the teeth of these three components are aligned, and a compressive force is applied between the lower spine and upper spine, the two units enter a coupled state, rotating together. Conversely, when the teeth are not aligned and no external force is applied, the two units are decoupled, allowing them to rotate independently. Taking the three-unit system in Fig.~\ref{fig:components_locking}E as an example, if no force is applied to any tendon ring, all three units remain decoupled. When force is applied to the tendon ring of Unit 1, Units 1 and 2 become coupled. If force is applied to both tendon rings, all three units become coupled. Similarly, when force is applied only to the tendon ring of Unit 2, Units 2 and 3 become coupled.

\begin{figure}[!t] 
    \centering
    \includegraphics[width=\linewidth]{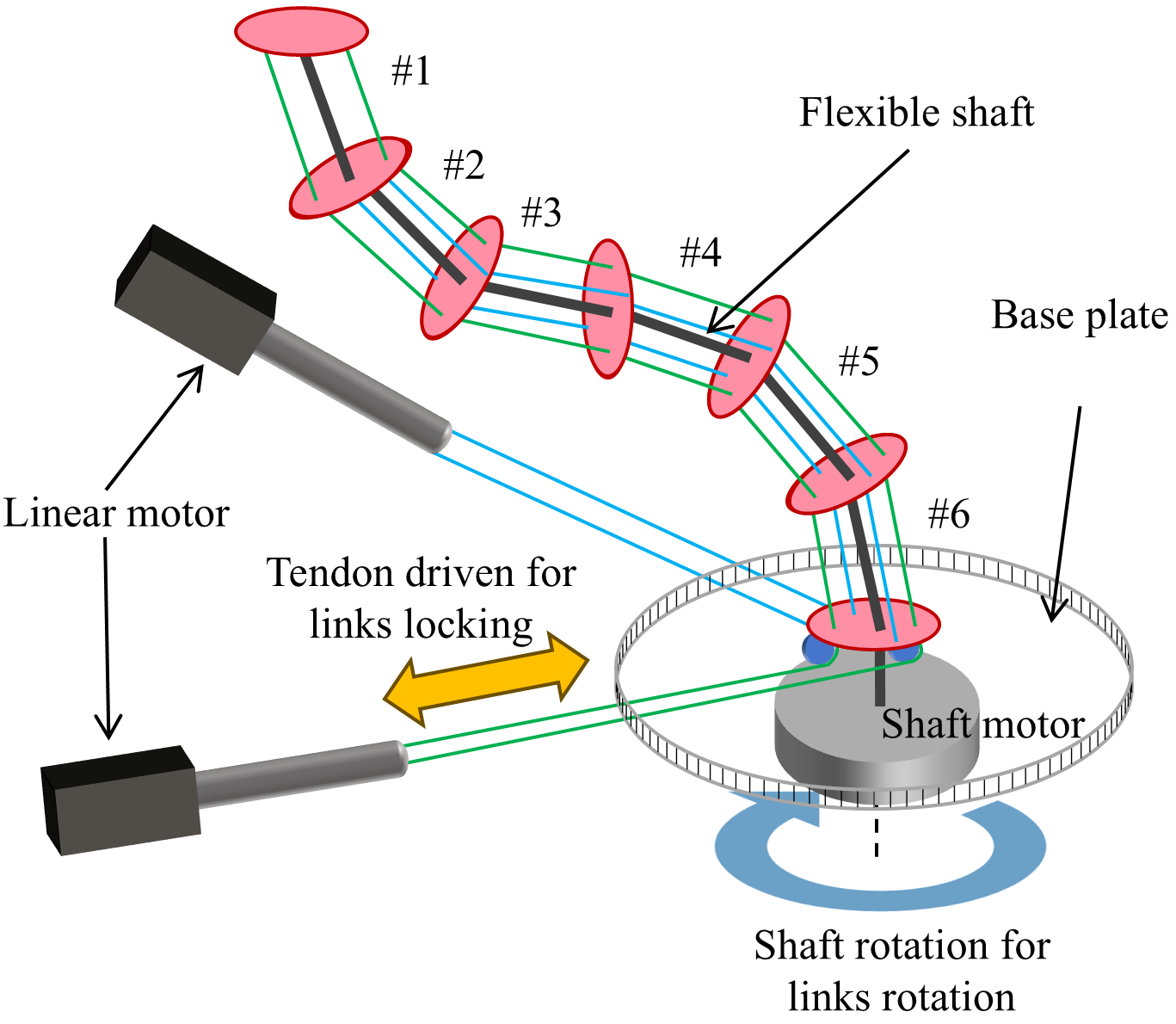}
    \caption{Illustration of the actuation principle. Tendons are routed through multiple PLC units. Each tendon connects to one linear motor. By pulling and loosening the tendon, linear motor controls the stiffness of one PLC unit. The rotation of PLC robot is controlled by shaft motor.
    }
    \label{fig:tendon}
\end{figure}

\begin{figure*}[!t] 
    \centering
    \includegraphics[width=\linewidth]{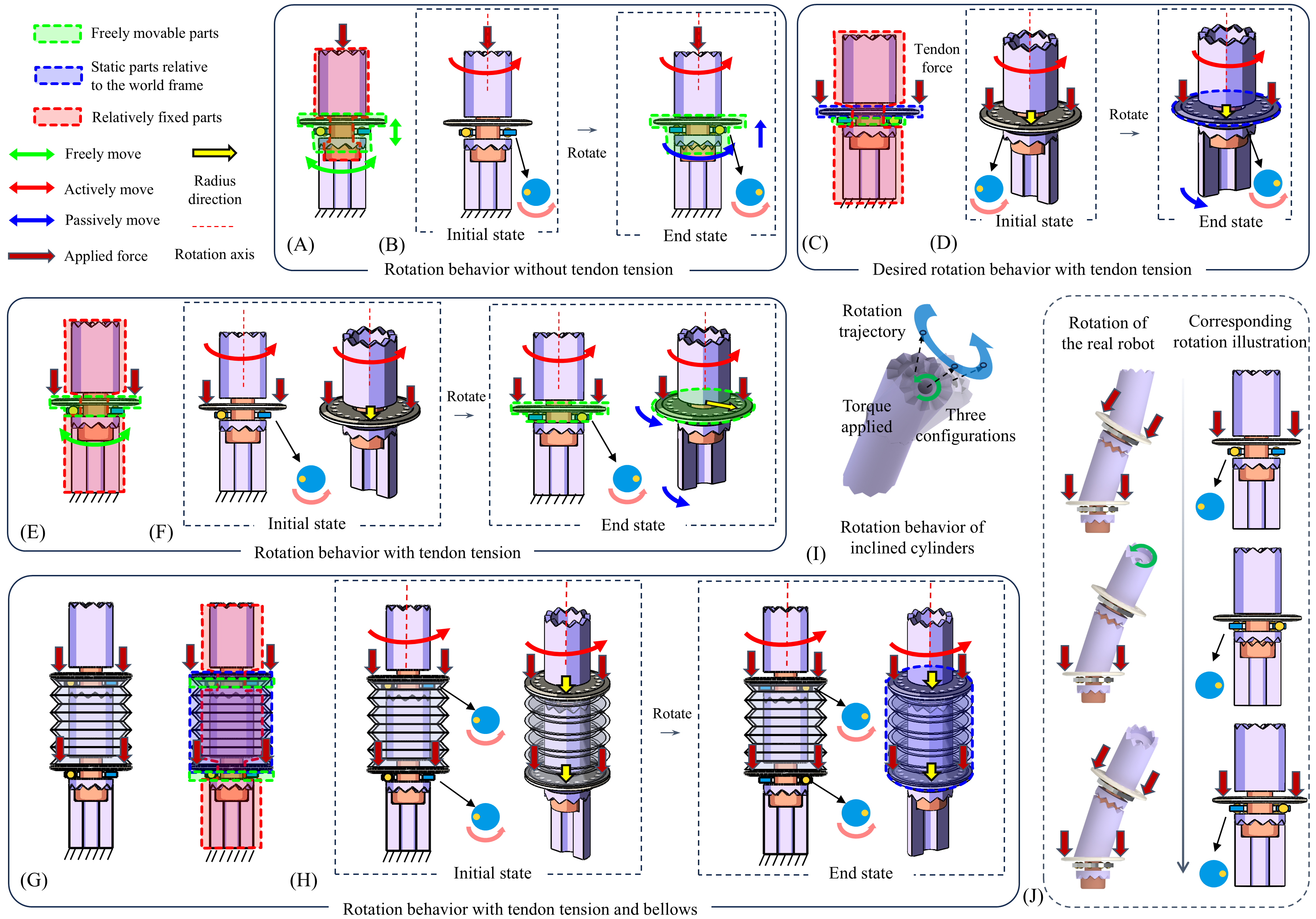}
    \caption{Principles of degrees of freedom. This explains how each serially connected PLC unit rotates under different structural configurations and tendon tensions, thereby illustrating how the PLC robot moves under different actuation conditions. (A) The PLC robot with force only applied to the spine structrure. (B) State change of the robot by applying external torque to the spine. (C) The PLC robot with force only applied by tendon. (D) Desired state change of the robot by applying external torque to the spine. Yellow arrow shows the rotation of the tendon ring. (E) The PLC robot with ony the tendon force. (F) The actual state change of the robot by applying torque on the spine. Yellow arrow shows the tendon ring rotates accordingly. (G) The PLC robot wearing the origami skin. (H) The state change of the PLC robot. Yellow arrow indicates the tendon ring doesn't rotate. (I) The rotation behavior of inclined modulars. Torque is only applied at the end of the serial modular. (J) The rotation behavior of the PLC robot. }
    \label{fig:dofmechanism}
\end{figure*}

\subsubsection{Actuation Principles}

The actuation of the PLC robot is categorized into two types: motion actuation and variable stiffness actuation. The variable stiffness actuation, as illustrated in Figs.~\ref{fig:components_locking}E and \ref{fig:tendon}, is achieved through tendon. The two ends of the tendon are tied at the tendon ring, exerting pressure on the tendon ring. The tendons are symmetrically distributed around the tendon ring to minimize potential twisting and ensure counterbalancing. The closed-loop end of the tendon displaces under applied tension, thereby regulating the stiffness of the corresponding unit.

The rotational motion actuation of the PLC robot is driven by a single flexible shaft, as depicted in Fig.~\ref{fig:tendon}. The flexible shaft passes through the passage of all units and is ultimately connected to the upper spine of the end unit. This configuration allows torque to be applied at the upper spine of the end unit, as indicated by the green arrow in Fig.~\ref{fig:dofmechanism}I. Notably, one PLC robot requires only one flexible shaft. By implementing timing control, we coordinate the rotation of all units. Further details regarding this mechanism are provided in the section on degrees of freedom principles.

\subsubsection{Principles of Degrees of Freedom}
The degrees of freedom can be divided into five behaviors based on Fig. \ref{fig:dofmechanism}.

\textbf{Behavior 1}: Rotation without tendon tension, as shown in Fig. \ref{fig:dofmechanism}A. We assume that an external force is applied to the upper spine of module 1, causing the lower spine of module 1 to connect tightly with the upper spine of module 2. In this case, components \#2 to \#4 of module 1 can move freely along the spine, including up and down and rotational motions. We consider module 1 as the end module and module 2 as the base module. As shown in Fig. \ref{fig:dofmechanism}B, in the initial state, when the end module is subjected to torque, the spine of module 1 rotates synchronously. However, since module 2 is fixed, the locking ring of module 1 is passively pushed upwards. At this point, the spine of module 1 rotates without actuating module 2, while the locking ring disengages and leads to undesired rotation of components \#2 and \#3.

\textbf{Behavior 2}: Rotation with tendon tension (Figs. \ref{fig:dofmechanism}E and \ref{fig:dofmechanism}F). When tension is applied to the tendon ring of module 1, a significant change occurs: the relatively fixed modules now extend from the spine of module 2 (Fig. \ref{fig:dofmechanism}A) to include the locking ring of module 1. Under pressure, the locking ring engages downwards, consolidating components \#4 to \#6 (Fig. \ref{fig:dofmechanism}E). As shown in Fig. \ref{fig:dofmechanism}F, in the initial state, when the end module is subjected to torque, the modules engaged with each other rotate synchronously. Therefore, the rotation of module 1 drives the rotation of module 2, as shown in the end state. On the other hand, components \#2 and \#3 of module 1 can still rotate freely. The bearing (\#3) here reduces the friction between \#2 and \#4; otherwise, the modules would rotate the tendon ring during rotation. However, the bearing only reduces friction, so the tendon ring still generates some rotation in this situation.

\textbf{Behavior 3}: Ideal rotation with tendon tension. The ideal rotation should involve the spines of the modules rotating while the bearing reduces rotational friction, and the tendon ring does not rotate. Since the rotation of the tendon ring disrupts the tendon’s routing between modules, severe disruptions can render the PLC robot inoperative.

\textbf{Behavior 4}: Addressing tendon ring rotation issues by introducing bellows, which is the structural design of our PLC robot in practical applications. As shown in Fig. \ref{fig:dofmechanism}G, the bellow connects the tendon rings of adjacent modules, forming a unified structure. The tendon ring of the last module connects to the base plate. Tendon ring rotation is caused by friction, tension, and external forces resulting in twists, while the bellows effectively resist these twists by serially connecting each tendon ring, as shown in Fig. \ref{fig:dofmechanism}H.

\textbf{Behavior 5}: Rotational motion of the inclined module (Fig. \ref{fig:dofmechanism}J). Similar to Fig. \ref{fig:dofmechanism}H, tension exists in both modules in the initial state, and the corresponding state of the linear module is also shown on the right side of the figure. When no tension is applied to the end module and torque is applied, the end module can rotate freely within 360 degrees. Finally, when the end module rotates to the desired discrete position, tension is applied to the end module again to lock the two modules together.
\subsection{Variable Stiffness Mechanism}

The stiffness of a PLC unit has two states: firmed stiffness and loosening stiffness.

\textbf{Firmed stiffness} occurs when the tension in the tendon is sufficiently high, causing the spines of multiple units to behave as a rigidly connected structure. In other words, firmed stiffness represents the maximum achievable stiffness of the PLC unit, where the structure can be considered a continuous, rigidly bent rod. Fig.~\ref{fig:components_locking}E illustrates the transition to firmed stiffness under high tension. Initially, when no tension is applied, the spines of Unit 1 remain independent. As high tension is applied to Unit 1, its spine becomes rigidly connected to that of Unit 2. Subsequently, when high tension is applied to Unit 2, the spines of Units 1, 2, and 3 all become rigidly connected, forming a single rigid structure. When the tension in Unit 1 is released, the connection between Units 1 and 2 is lost, allowing relative rotation between them.

\textbf{loosening stiffness} refers to stiffness values that fall between the minimum and maximum stiffness states. When the tendon is fully relaxed, the connections between units become highly flexible, resulting in minimal stiffness (only stiffness from origami skin). Conversely, under high tension, the stiffness reaches its firmed state. Therefore, loosening stiffness is directly dependent on the tendon tension: as the tension increases, the stiffness also increases, ultimately approaching its maximum value. A more detailed analysis of this relationship is provided in the theoretical analysis section.

\subsection{Programmable Behavior of the PLC Robot}

\begin{figure}[!t] 
    \centering
    \includegraphics[width=\linewidth]{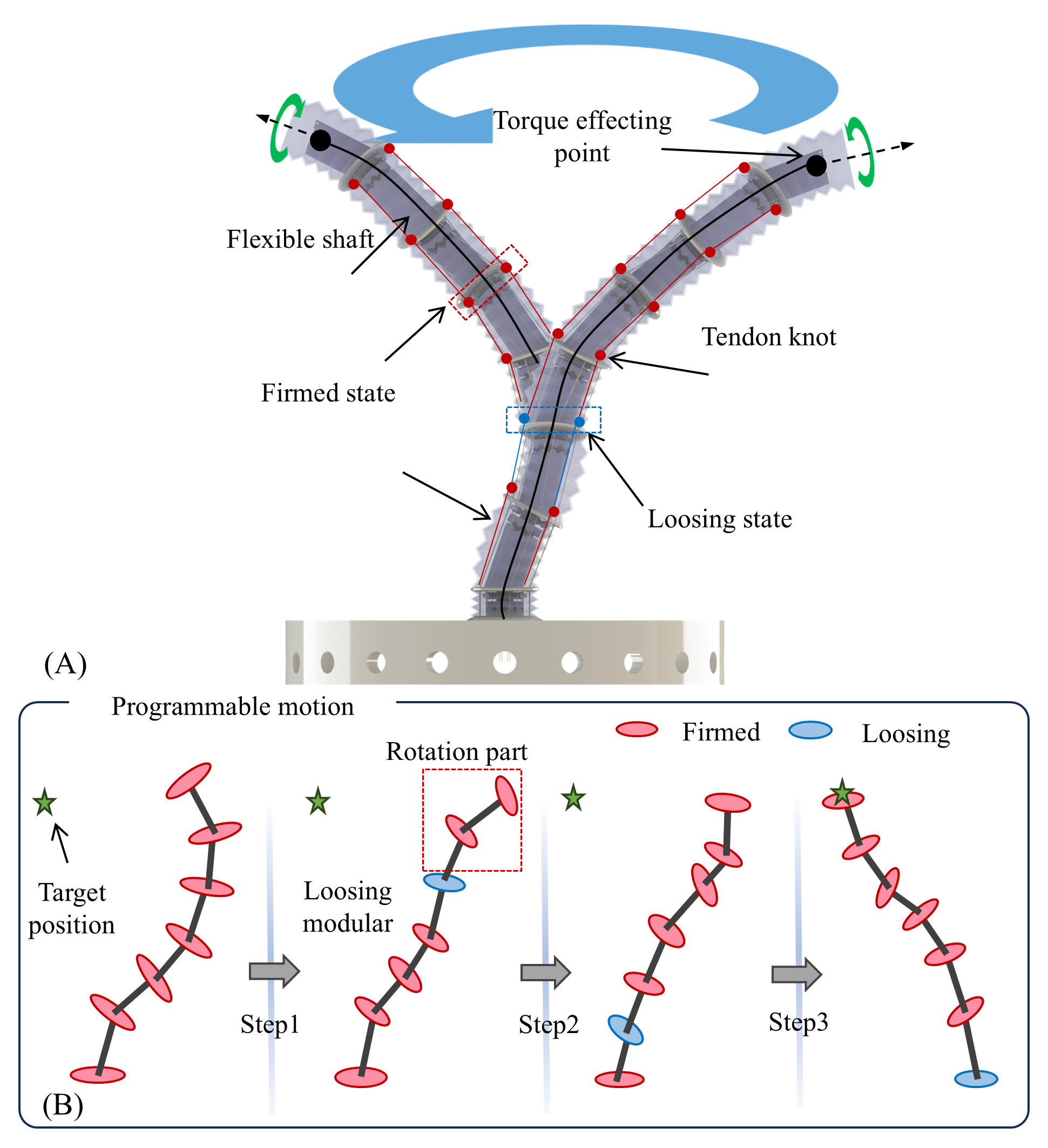}
    \caption{
        Programmable behavior of the PLC robot. (A) The rotation behavior of a multi-segments PLC robot. (B) Point-reaching task accomplished by programmably adjusting the stiffness and joint angles of each unit.
    }
    \label{fig:programmingmotion}
\end{figure}

Consider a PLC robot composed of seven modules, each with independently controlled tendons at its tendon ring, as shown in Fig. \ref{fig:programmingmotion}A. The central black line represents the flexible shaft running through the robot, which is connected to the end module. On each module's tendon ring, two points represent the two knots of a tendon. The tendons controlling each module pass through different holes in the tendon ring, allowing the six tendon rings to be individually controlled by six motors, which can switch between unlock and lock states. In a sequential programming behavior, only one module is set to the loosening state at a time, while the remaining modules stay in the locking state. As shown in Fig. \ref{fig:programmingmotion}A, with one module in the loosening state, the entire upper section can rotate under the torque applied by the flexible shaft.

To illustrate, the example of an end-point reaching task demonstrates a complete programming behavior, as shown in Fig. \ref{fig:programmingmotion}B. For a desired end-point, we calculate the robot's configuration space, i.e., the angle of each module. Then, we sequentially rotate each corresponding joint to the desired angle, ultimately achieving the end-point.

\section{Kinematics Analysis}

This section models the PLC robot kinematics and details a motion planning methodology to achieve a desired end-effector position.  Given the discrete nature of the joint pose space, a k-d tree structure is employed to facilitate efficient inverse-kinematics search operations. Initially, the kinematic properties of the PLC robot are characterized across the following three spaces:
 
\textbf{Actuator Space} captures the motion of the motors. It is represented as $\boldsymbol{a} = \begin{bmatrix} m & \zeta_1 & \dots & \zeta_n \end{bmatrix}^{\intercal} $, where $\zeta_1,\dots,\zeta_n\in\mathbb{R}$ denotes the extension stages of the $n$ linear tendons, $m \in \mathbb{R}$ represents the motor position. Smaller value of $\zeta_i$ indicates that the tendon $i$ is tighter, making the teeth between segments $i$ and $i+1$ clutched more firmly with each other. On the contrary, larger value of $\zeta_i$ indicates that the tendon $i$ is looser, making the teeth between segments $i$ and $i+1$ less firmly clutched to each other.

\begin{figure}[!t] 
    \centering
    \includegraphics[width=\linewidth]{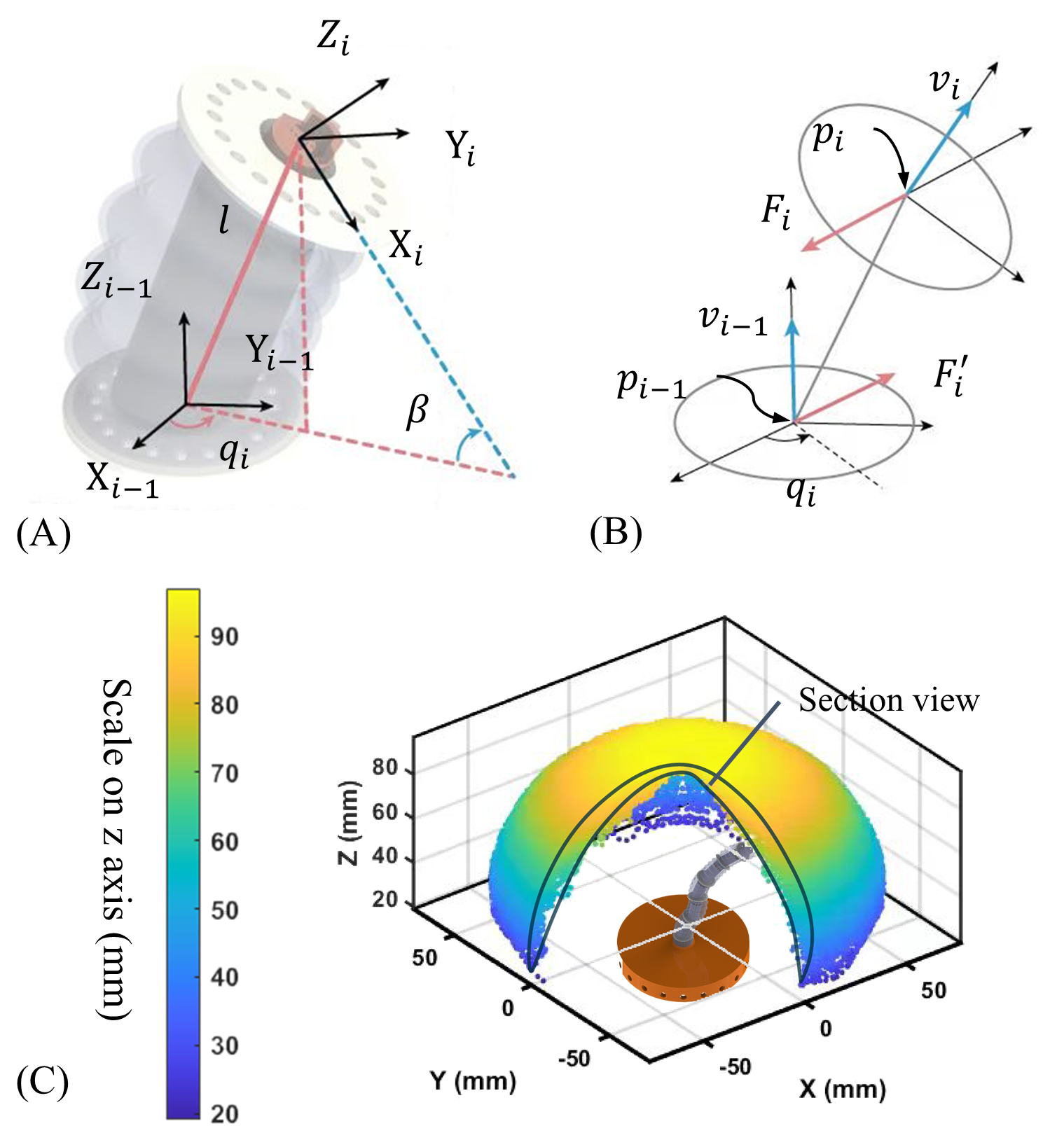}
    \caption{
       Kinematics of the PLC robot. (A) Modular coordinates. (B) Abstracted modular structure. (C) Working space of a 5-segments PLC robot, which is discrete. The section view indicates surface thickness of the working area.
    }
    \label{fig:kinematics}
\end{figure}

\textbf{Configuration Space} describes the rotation angles of each modular of the robot joints as $\boldsymbol{q} = \begin{bmatrix} q_1 & \dots & q_n \end{bmatrix}^{\intercal} \in \mathbb{R}^n$, where $q_i$ is the rotation angle around the Z-axis of the $(i-1)$-th segment (Fig. \ref{fig:kinematics}A).

\textbf{Task Space} represents the position and orientation of the robot's end-effector ($\boldsymbol{R}_e, \boldsymbol{p}_e$) and the tool-tip of the swab ($\boldsymbol{R}_t, \boldsymbol{p}_t$) in $SE(3)$ \cite{lynch2017modern}, both of which are in cartesian space.

\subsubsection{Single-segment kinematics}

As illustrated in Fig. \ref{fig:kinematics}B, we derive the kinematics for transforming parameters from the configuration space $\boldsymbol{q}$ to the task space $\{\mathbf{x}\}$. The position of the distal end of the $i$-th segment, ${}^{i-1}\mathbf{p}_i = \begin{bmatrix} {}^{i-1} x_i & {}^{i-1} y_i & {}^{i-1} z_i \end{bmatrix}^{\intercal}$, is given by:
\begin{equation}
    \begin{aligned} 
        {}^{i-1} x_i &= \frac{L}{\beta}(1 - \cos{\beta}) \cos{q_i}\\
        {}^{i-1} y_i &= \frac{L}{\beta}(1 - \cos{\beta}) \sin{q_i}\\ 
        {}^{i-1} z_i &= \frac{L}{\beta} \sin{\beta}
    \end{aligned}
\end{equation}
where $L$ is the length of the center curve, and $\beta$ is the bending angle of each segment, both of which are constant and customizable. The rotation matrix, ${}^{i-1} \mathbf{R}_i = RotZ(q_i) \cdot RotY(\beta)$, is given by
\begin{equation}
    {}^{i-1} \mathbf{R}_i = 
    \begin{bmatrix}
        c_{q_i}c_{\beta} & -s_{q_i} & c_{q_i}s_{\beta} \\
        s_{q_i}c_{\beta} & c_{q_i} & s_{q_i}s_{\beta} \\
        -s_{\beta} & 0 & c_{\beta}
    \end{bmatrix}
\end{equation}
where $s_{q_i}$ and $c_{q_i}$ abbreviate $\sin{q_i}$ and $\cos{q_i}$, respectively.

\begin{figure}[!t] 
    \centering
    \includegraphics[width=\linewidth]{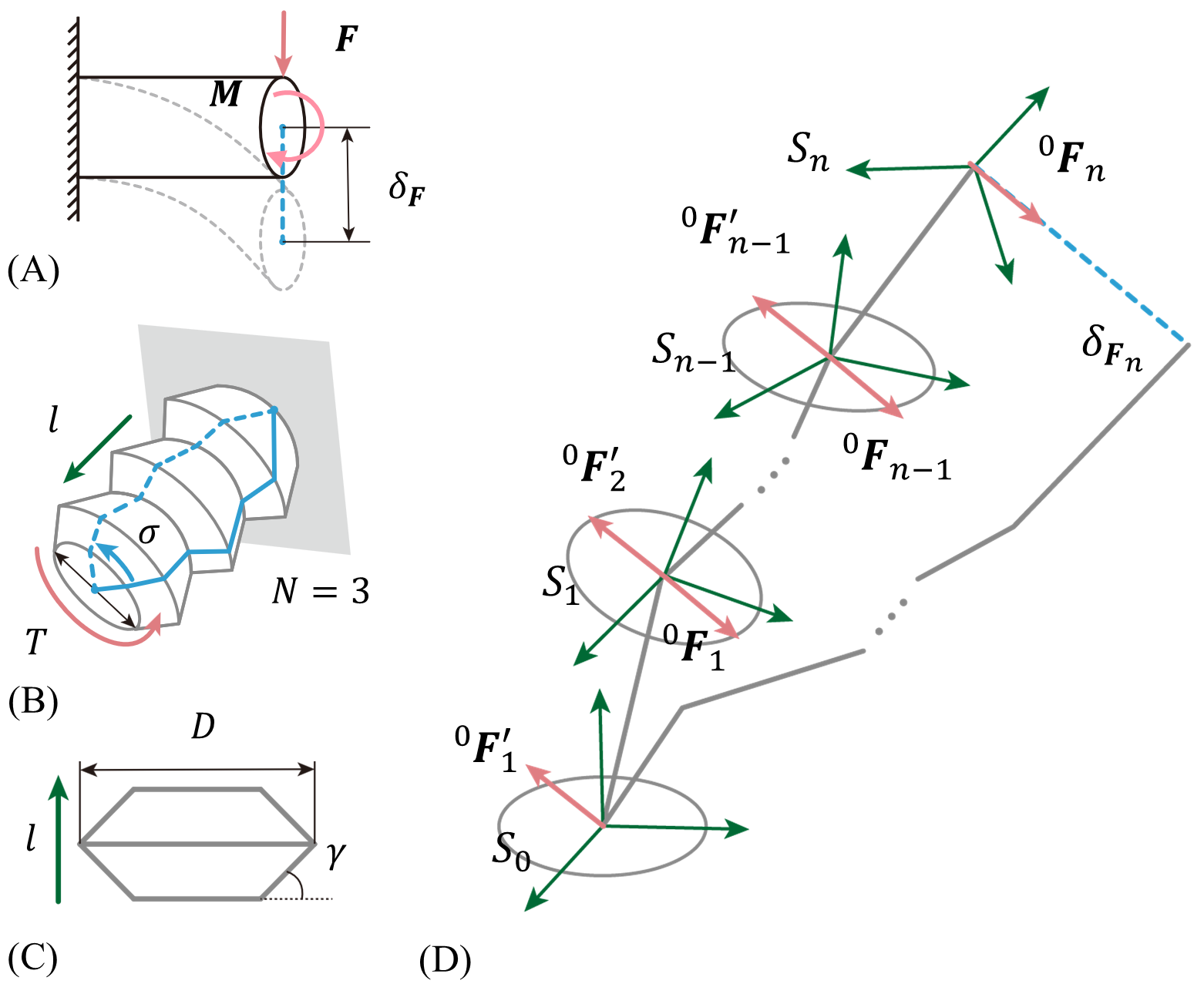}
    \caption{
        Firmed stiffness modeling. (A) The bending of beam. (B) The twising of the origami skin. (C) The origami cross section view. (D) The deformation illustration of an n-segments PLC robot.
    }
    \label{fig:lockingstiffness}
\end{figure}

\subsubsection{Multi-segment Kinematics}

By combining the transformations of each adjacent segment, we derive the multi-segment kinematics as:
\begin{equation}
    {}^{b} \mathbf{T}_{e} = {}^{b} \mathbf{T}_{1}(q_1) ... {}^{i-1} \mathbf{T}_{i}(q_i) ... {}^{n-1} \mathbf{T}_{n}(q_n)
\end{equation}
where $n$ represents the number of modulars, $T$ is the homogeneous transformation matrix,

\begin{equation}
    {}^{i-1} \mathbf{T}_{i} 
    =
    \begin{bmatrix} 
        {}^{i-1} \mathbf{R}_{i} & {}^{i-1} \mathbf{p}_{i} \\
        \mathbf{0} & 1
    \end{bmatrix}
\end{equation}
where $\mathbf{0}$ is a zero vector.

Considering a tool is attached at position ${}^{e}\mathbf{p}_t = \begin{bmatrix} x_t & y_t & z_t \end{bmatrix}^{\intercal}$ relative to the end-effector, the tool-tip pose relative to the base frame is given by:
\begin{equation}
    {}^{b} \mathbf{p}_t = {}^{b} \mathbf{R}_e {}^{e} \mathbf{p}_t + {}^{b} \mathbf{p}_e.
\end{equation}

The working space composed of discrete reachable points are shown in the Fig. \ref{fig:kinematics}C. The density of points can be intuitively represented using omnivariance as shown in the Appendix \ref{appenA}.
\subsubsection{Search-based Inverse Kinematics}

Due to the high redundancy of the robot, we use a k-nearest neighbors (k-NN) algorithm \cite{abbasifard2014survey} on a k-d tree \cite{bentley1975multidimensional} to solve the inverse kinematics problem. The following algorithms outline the process.

First, we construct the k-d tree by exhaustively exploring the robot's state space using Depth-First Search (DFS) \cite{cormen2022introduction}, as shown in Algorithm 1. Because of the robot's structural redundancy and symmetry, multiple configurations can map to the same task space configuration. To handle this, we use a map structure to store all configurations that correspond to the same task space configuration. This precomputation, while time-consuming, needs to be done only once.

During each search, the k-NN algorithm is employed to find the nearest node in the k-d tree, as outlined in Algorithm 2. This allows us to find the configuration closest to a reference configuration.

\begin{algorithm}[!t]
\caption{Build k-d Tree with Discrete Joint Angles}
\begin{algorithmic}[1]
\REQUIRE Joint size $n$, Pre-curved bending angle $\beta$, Segment curve length $L$
\ENSURE k-d tree $kd\_tree$ to save the task configurations, task to configuration space mapping $task\_to\_joint\_map$
\STATE $discrete\_angles \leftarrow [0, 36, 72, ..., 324]$
\STATE $task\_configuration\_list \leftarrow []$
\STATE $task\_to\_joint\_map \leftarrow []$
\STATE $stack \leftarrow []$
\STATE $stack.push(\{state = \bm 0_{n \times 1}, joint\_id = 1\})$
\WHILE{not $stack$.empty()}
    \STATE $current\_node \leftarrow stack.\text{pop}()$
    \STATE $\bm \theta^{current} \leftarrow current\_node.state$
    \STATE $idx \leftarrow current\_node.joint\_id$
    \IF{$idx > n$}
        \STATE $\bm T \leftarrow$ multi\_segment\_fk($\bm \theta^{current}, \beta, L$)
        \STATE $key \leftarrow$ generate\_key\_from\_position($\bm T.position$) 
        \IF{$key$ exists in $task\_to\_joint\_map$}
            \STATE $task\_to\_joint\_map.\text{append}(key, \bm \theta^{current})$
        \ELSE
            \STATE $task\_to\_joint\_map.\text{insert}(key, \bm \theta^{current})$
            \STATE $task\_configuration\_list.\text{append}(\bm T.position)$
        \ENDIF
    \ELSE
        \FOR{each $angle$ in $discrete\_angles$}
            \STATE $\bm \theta^{next} \leftarrow \bm \theta^{current}$
            \STATE $\bm \theta^{next}_i \leftarrow angle$
            \STATE $stack.push(\{\bm \theta_{next}, idx + 1\})$
        \ENDFOR
    \ENDIF
\ENDWHILE
\STATE $kd\_tree \leftarrow \text{generate\_kd\_tree}(task\_configuration\_list)$
\end{algorithmic}
\end{algorithm}

\begin{algorithm}[!t]
\caption{Inverse Kinematics using k-NN Tree Search}
\begin{algorithmic}[1]
\REQUIRE Desired end-effector position ${}^b \bm p_e^d$, k-d tree $kd\_tree$, mapping $task\_to\_joint\_map$, Configuration reference $\bm \theta^\star$
\ENSURE Joint configuration $\bm \theta$ achieve ${}^b \bm p_e^d$ with closest to $\bm \theta^\star$
\STATE $idx \leftarrow$ knnsearch($kd\_tree$, ${}^b \bm p_e^d$)
\STATE ${}^b \bm p_e^{nearest} \leftarrow kd\_tree(idx)$
\STATE $key \leftarrow$ generate\_key\_from\_position($position$)
\STATE $\bm \theta^{candidates} \leftarrow task\_to\_joint\_map.\text{find}(key)$
\STATE $\bm \theta \leftarrow \bm 0_{n \times 1}$
\STATE $min\_dist \leftarrow \infty$
\FOR{each $\bm \theta_i$ in $\bm \theta^{candidates}$}
    \STATE $dist \leftarrow \Vert \bm \theta - \bm \theta^\star \Vert_2$
    \IF{$min\_dist > dist$}
        \STATE $min\_dist \leftarrow dist$
        \STATE $\bm \theta \leftarrow \bm \theta_i$
    \ENDIF
\ENDFOR
\end{algorithmic}
\label{algo:inverse}
\end{algorithm}

\section{Stiffness Modeling}
\label{sec:stiffness modeling}

This section presents a theoretical analysis of the stiffness control mechanism in the PLC robot. As previously discussed in Section \ref{sec:design}, the deformation of the PLC robot under an external force arises from two primary sources: trunk deformation and tendon deformation. Section \ref{subsec:locking stiffness} elaborates on trunk deformation, where the robot can be modeled as a single solid body, and its deformation adheres to the principles of continuum mechanics. Section \ref{subsec:unlocked stiffness} details tendon deformation, which occurs when the locking tendons begin to stretch under load.  Finally, an analysis of torsional stiffness is provided to demonstrate how the origami skin effectively prevents twisting displacement of the locking ring during deformation.

\subsection{Firmed Stiffness}
\label{subsec:locking stiffness}

\begin{figure}[!t]\centering
	\includegraphics[width=\linewidth]{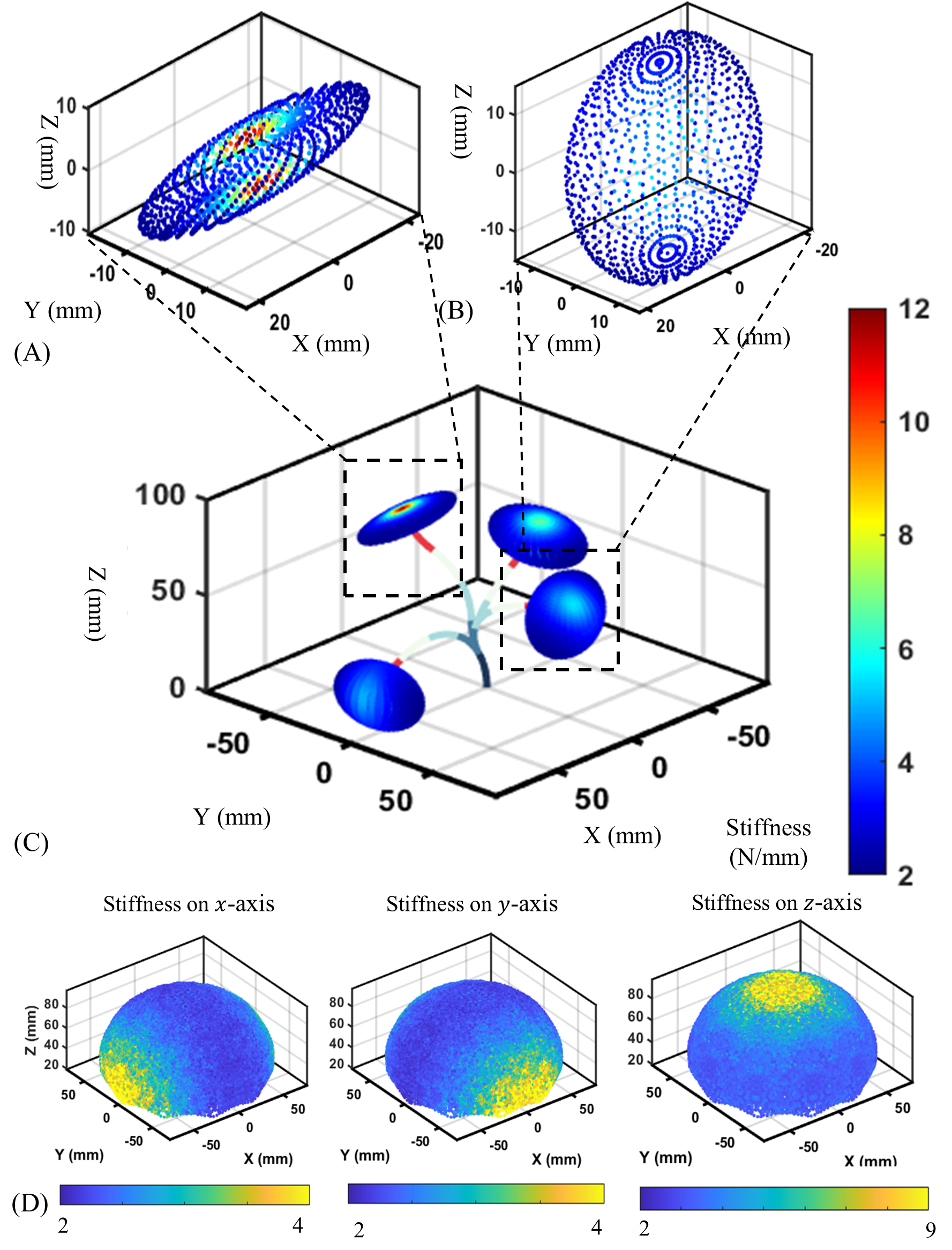} 
	\caption{(A) Zoomed-in stiffness distribution of configuration one. (B) Zoomed-in stiffness distribution of configuration two. (C) The stiffness distribution in four different configurations. (D) The stiffness in x, y, and z directions in working space for a 5-segments PLC robot.}
	\label{fig:omnistiffness}
\end{figure}

The maximum attainable stiffness of the PLC robot within an infinitesimal local domain is termed "firmed stiffness". This condition arises when trunk deformation is the sole contributor to the overall deformation of the PLC robot, occurring when the external force is insufficient to overcome the tendon tension. As detailed in Section \ref{subsec:motion mechanism}, the locking ring remains firmly engaged under these circumstances, effectively rendering the PLC robot a single, tube-shaped solid structure. Consequently, the firmed stiffness can be analyzed using classical continuum elastic mechanics.

Each segment of the PLC robot can be viewed as a elastic rod as shown in Fig. \ref{fig:lockingstiffness}A, whose strain energy consists of a bending and an axial component.
\begin{equation}
    U_i = \prescript{i}{} U_b+\prescript{i}{} U_n,\quad i=1,\cdots,n
    \label{eq:sum_strain}
\end{equation}

$\prescript{i}{}U_b$ is the elastic potential energy stored in tangential deformation caused by the bending moment component distribution $\mathbf{M}_i(r)$, which is a function of $r$, the distance to the segment base. $\prescript{i}{} U_n$ is the elastic potential energy in normal deformations raised by the axial pressing force component distribution $\mathbf{N}_i(r)$, which is also a function of $r$. Both $\mathbf{M}_i(r)$ and $\mathbf{N}_i(r)$ are sourced from the external force $\mathbf{F}_i$ on the segment. The mathematical forms of $\prescript{i}{}U_b$ and $\prescript{i}{} U_n$ can be written as the following.
\begin{equation}
\begin{aligned}
    \prescript{i}{}U_b &= \int \frac{|\mathbf{M}_i(r)|^2}{2EI}dr \\
    \prescript{i}{}U_n &= \int \frac{|\mathbf{N}_i(r)|^2}{2EA}dr
\end{aligned}
\label{eq:bending_axial}
\end{equation}
where $E$ is the young's modulus of the material, $I$ is the polar moment of inertia of the cross-section and $A$ is the cross-sectional area. Since the origami skin, as shown in Fig. \ref{fig:lockingstiffness}B, has very high twisting stiffness and negligible pressing and shearing stiffness, the twisting component is dropped in the strain energy (\ref{eq:sum_strain}). The bending torque $\mathbf{M}_i(r)$ and axial loading $\mathbf{N}_i(r)$ of the $i$-th segment, as shown in Fig. \ref{fig:lockingstiffness}A, are
\begin{equation}
    \mathbf{M}_i(r)=(-r\mathbf{v}_i)\times {\mathbf{F}_i}
\label{eq:rotbending}
\end{equation}
\begin{equation}
    \mathbf{N}_i(r)=(\mathbf{v}_i \cdot \mathbf{F}_i)\mathbf{v}_i
\label{eq:normalforce}
\end{equation}
where $\mathbf{v}_i$ is the axial unit vector at the base point $\boldsymbol{p}_i$ in the world frame. This leads to a closed form of (\ref{eq:bending_axial})

\begin{equation}
\begin{aligned}
    \prescript{i}{}U_b &= \frac{|\mathbf{v}_i\times\mathbf{F}_i|^2L^3}{6EI}=\frac{[|\mathbf{F}_i|^2-(\mathbf{v}_i\cdot\mathbf{F}_i)^2]L^3}{6EI} \\
    \prescript{i}{}U_n &= \frac{(\mathbf{v}_i\cdot\mathbf{F}_i)^2L}{2EA}
\end{aligned}
\label{eq:bending_axial_closed}
\end{equation}

Due to the light weight of PLC robot, the external force $\boldsymbol{F}_i$ on each segment also follows the transmission rule $ \boldsymbol{F}_i=-\boldsymbol{F}_i, $ and $\boldsymbol{F}_{i-1}=-\boldsymbol{F}_i$. This indicates $\mathbf{F}_1=\mathbf{F}_2=\cdots=\mathbf{F}_n$.

Based on the Castigliano's method, we attain the elastic displacement $\boldsymbol{\delta}_i$ of each segment from the gradient of $U_i$ against the general force, which is $\mathbf{F}_i$ in our case.
\begin{equation}
    \boldsymbol{\delta}_i=\nabla_{\mathbf{F}_i}U_i
\label{eq:castigliano}
\end{equation}

The overall deformation on the PLC robot end effector is the summed deformation over all segments, $\boldsymbol{\delta}_n=\sum\boldsymbol{\delta}_i$, as indicated in Fig. \ref{fig:lockingstiffness}D. Plugging in (\ref{eq:bending_axial_closed}) and (\ref{eq:castigliano}), and apply the transmission rule, the following can be inferred
\begin{equation}
\begin{aligned}
    \boldsymbol{\delta}_n&=\mathbf{K}_n\mathbf{F}_n \\
    \mathbf{K}_n&=\sum_{i=1}^n\Big[\frac{L(\mathbf{v}_i\bigotimes\mathbf{v}_i)}{EA}+\frac{L^3}{3EI}(\mathbf{I}-\mathbf{v}_i\bigotimes\mathbf{v}_i)\Big]
\end{aligned}
\label{eq:displace_predict}
\end{equation}
where $\bigotimes$ is the outer product of two vectors.

The stiffness matrix, $\mathbf{K}_n$, exhibits anisotropic characteristics, with its elements dependent on the axial unit vector of each segment within the PLC robot. This property is illustrated in Figure \ref{fig:omnistiffness}, which depicts the directional stiffness in four distinct configurations. In each configuration, a 50 N external force is applied to the end-effector along sampled directions in spherical coordinates. The predicted end-effector displacements, calculated using Equation (\ref{eq:displace_predict}), are presented in Figure \ref{fig:omnistiffness}C, visually demonstrating the anisotropic nature of the firmed stiffness. The color of each point represents the stiffness associated with the corresponding deformation, computed as $|\boldsymbol{\delta}_n|/|\mathbf{F}_n|$. Figure \ref{fig:omnistiffness}D further emphasizes this anisotropy by displaying the stiffness along the x, y, and z directions for all $10^5$ configurations of the 5-segment PLC robot model.

\subsection{loosening Stiffness}
\label{subsec:unlocked stiffness}

When the external force exceeds a critical threshold, overcoming the tendon tension, the locking ring may disengage, precluding the modeling of the PLC robot as a single, unified structure, as described in Section \ref{subsec:locking stiffness}. In this scenario, further deformation is primarily attributed to tendon elongation. This section examines the conditions under which locking ring detachment occurs and establishes a model to characterize the force-deformation relationship in this regime. The stiffness of the PLC robot following locking ring detachment is referred to as "loosening stiffness."

\begin{figure}[!t] 
    \centering
    \includegraphics[width=\linewidth]{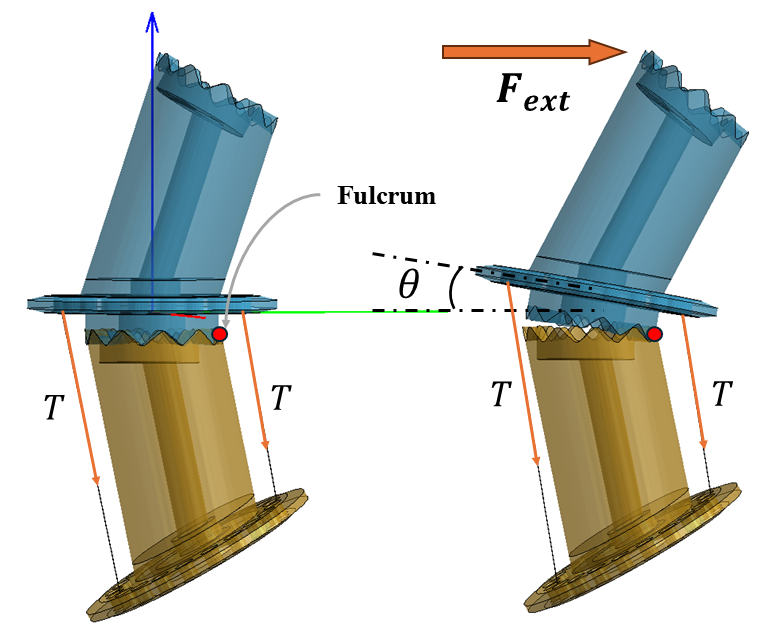}
    \caption{Schematic diagram when the locking ring detaches because of external force. The upper and lower segments are respectively colored blue and yellow. When there is no or small external force, the locking ring is firmly detached to the lower segment under the tendon tension as shown on the left, which is the home state. When an considerable external force is applied on the upper segment, the locking ring will detach up to an small open angle $\theta$, as shown on the right. PLC robot shows the unlock stiffness where further deformation is mainly contributed by tendon stretch.}
    \label{fig:bending_stiffness_modelling}
\end{figure}

Consider two adjacent segments of the PLC robot segment as shown in Fig. \ref{fig:bending_stiffness_modelling}, where tension $\mathbf{T}$ is applied to the tendon. An external force $\mathbf{F}_{ext}$ is applied to the top of the upper segment at position $\mathbf{p}_{ext}$. The detachment of the locking ring on the lower segment has a mechanism similar to a lever whose fulcrum $\mathbf{p}_{fu}$ is on the locking ring edge and along the direction of $\mathbf{F}_{ext}$. All vectors and coordinates are viewed in the base frame of the upper segment as shown in Fig. \ref{fig:bending_stiffness_modelling}, where the red, green and blue lines respectively denotes the $x$, $y$ and $z$ axis. $\mathbf{F}$ generates external torque
\begin{equation}
    \mathbf{\tau}_{ext}=(\mathbf{p}_{ext}-\mathbf{p}_{fu})\times\mathbf{F}_{ext}
\end{equation}

Section \ref{sec:design} has stated that tendon end points on the bearings are all distributed in a centrally symmetric manner, so if one tendon end-point on the upper segment bearing (the blue bearing in Fig. \ref{fig:bending_stiffness_modelling}) is $\mathbf{p}_e$, then the other tendon end point will be $-\mathbf{p}_e$. As such , the resistance torque by the tendon is
\begin{equation}
    \mathbf{\tau}_{res}=(\mathbf{p}_e-\mathbf{p}_{fu})\times\mathbf{T}_1+(-\mathbf{p}_e-\mathbf{p}_{fu})\times\mathbf{T}_2
\label{eq:tendon torque}
\end{equation}
$\mathbf{T}_1$ and $\mathbf{T}_2$ are tension vectors on the two tendons and they shares the same magnitude. When locking rings are at the home pose, $\mathbf{T}_1$ and $\mathbf{T}_2$ will also share the same direction, leading to $\mathbf{T}_1=\mathbf{T}_2=:\mathbf{T}$. So (\ref{eq:tendon torque}) can be simplfied to
\begin{equation}
    \mathbf{\tau}_{res}=-2\mathbf{p}_{fu}\times\mathbf{T}
\label{eq:tendon torque simplified}
\end{equation}

When $|\mathbf{\tau}_{ext}|>|\mathbf{\tau}_{res}|$, the locking ring teeth will open as indicated in Fig. \ref{fig:bending_stiffness_modelling}. This imposes conditions for the magnitude of $\mathbf{F}_{ext}$ for the PLC robot to leave firmed stiffness and enter loosening stiffness. Experiments will be conducted in section \ref{sec:experiments} to verify this model's accuracy.

Further increasing $\mathbf{F}_{ext}$ beyond the loosening threshold will lead to the stretching of the tendon. Since the tendon has significantly lower stiffness than the spine, it becomes the primary contributor to the deformation. Assuming tension $\mathbf{T}$ is insufficient for the tendon to enter the nonlinear zone, this model anticipates that the tension $\mathbf{T}$ will not considerably affect the overall loosening stiffness, even though it determines the external force threshold to transit from firmed stiffness to loosening stiffness. This prediction will also be supported in section \ref{sec:experiments}.

\subsection{Twisting Stiffness}
Although the continuum robot can increase its stiffness by pulling the cable, the antagonistic force, when not perfectly aligned within one surface, can produce a torque on the cross-section and thus generate undesirable twists. To analyze this twist and find out how well can this continuum robot resist the twist, we model the twisting stiffness for both the rigid links of the robot and soft shells of the robot.


\begin{figure}[!t] 
    \centering
    \includegraphics[width=\linewidth]{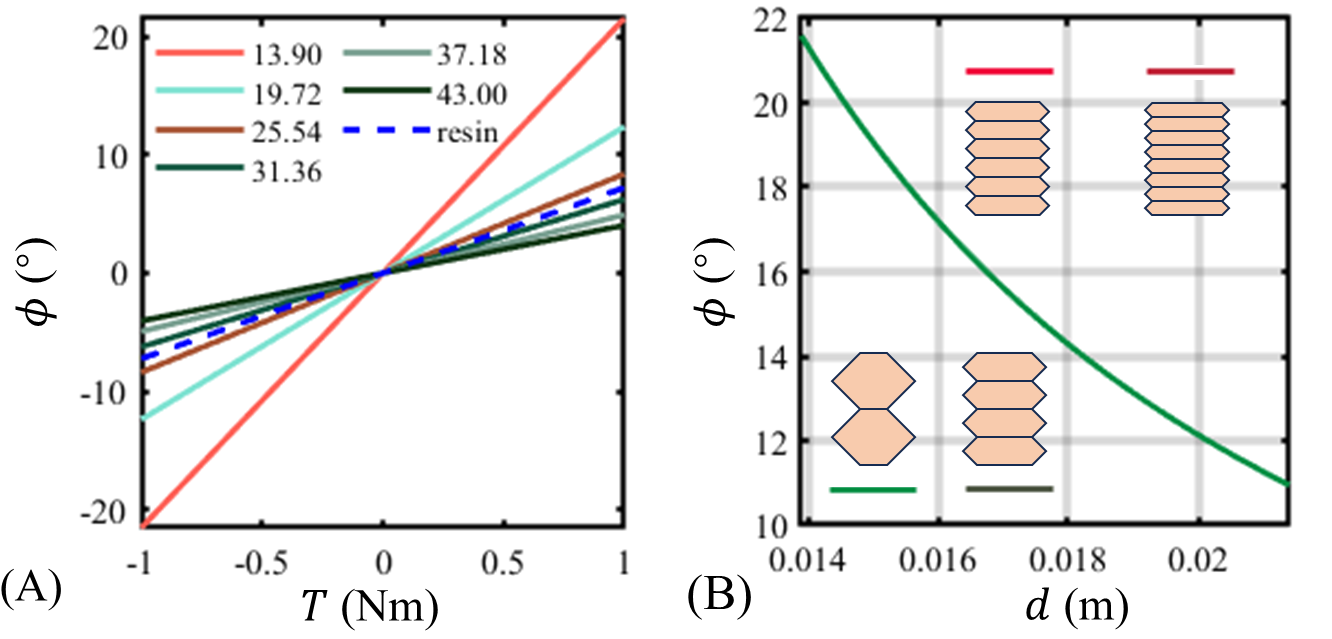}
    \caption{(A) Twisting stiffness comparison between origami skin and the resin spine structure. Different skin diameters are presented. (B) The twisting angle of the origami skin under $T = 1 Nm$. The x axis shows different diameter $d$ of the origami skin, and y axis is the twisting angle under the constant torque. Different convolutions (2, 4, 6, 8) are plotted in the figure, but they overlapped with each other. This indicates the number of convolutions does not influence the twisting stiffness.}
    \label{fig:experiment_setup}
\end{figure}

\textbf{Twisting stiffness of rigid structure} As shown in the Fig. \ref{fig:lockingstiffness}B, torsion on one segment is brought by the slight deviation of antagonistic cables and the shaft motor. As the segment is inclined, the torsion can be divided into axial torque and radial torque. The torsion formula, exploiting the relation between the axial torque and the twist at the cross-section (Fig. \ref{fig:lockingstiffness}C), is shown below,
\begin{equation}
    d \sigma=\frac{Tdl}{JG}
    \label{eq:torsion_formula}
\end{equation}
where $\sigma$ represents the twist angle, $T$ is the internal torque , $l$  is the length along the torque axis, $J$ is the polar moment of inertial at the cross section, $G$ is the modulus of rigidity. The polar moment of inertial of circle ring is $\frac{1}{2}\pi(c_2^4-c_1^4)$.

Given the outer diameter $c_2$ and inner diameter $c_1$, we attain the twist neglecting the slope
\begin{equation}
    \Delta \sigma = \frac{2l_0T}{\pi(c^4_2-c^4_1)G}
\end{equation}
where $l_0$ is the length of one segment.

\textbf{Twisting stiffness of origami skin} The twisting stiffness of rigid structure is enabled by pulling the locking cables such that the gears are engaged to compose one rigid body. However,  this twisting stiffness of segment $i$ becomes 0 in rotation mode, as the cables are released for its free movement. A soft bellow shell maintains a twisting resistance while not affecting the dimension and movement of the continuum robot.

The polar moment of inertia of the bellow shell is
\begin{equation}
    J_b(l)=\frac{1}{2}\pi\Big[\Big(\frac{D(l)}{2}\Big)^4-\Big(\frac{D(l)}{2}-t\Big)^4\Big]
\end{equation}
where $D(l)$ is the diameter at $l$, $t$ is the thickness of the shell. For one convolution, as the upper half is symmetrical to the lower half, we attain the twisting angle by
\begin{equation}
    \delta \sigma=\int^{\frac{h_0}{2}}_0 \frac{T}{J_b(l)G}dl
\end{equation}
where $h_0$ is the height of one convolution and can be expressed as $\frac{l_0}{N}$. The twisting angle of one segment is
\begin{equation}
    \Delta \sigma =2N \delta \sigma
\label{eq:delta_sigma_comp}
\end{equation}
where $N$ is the number of convolutions.

Through straightforward arithmetric computation, we will see that $\Delta\sigma$ only depends on the segment length $l_0$, inner diameter and outer diameter when the origami skin has a periodic zig-zag shape. Fig. \ref{fig:experiment_setup} illustrates this fact by plotting the $\Delta\sigma$ using numerical integration under different diameters and numbers of convolutions.


\section{Experimental Verification} 
\label{sec:experiments}

In this section, we present the experimental verification of the PLC robot's design and performance, focusing on its stiffness modulation and operational precision. We begin by outlining the experimental setup, including the configuration of the hardware and the calibration of the measurement instruments across experimental validations and demonstrations. Then, detailed experiments of the robot's stiffness characteristics—loosening and firmed—under various conditions is validated to compare with the theoretical results. This is followed by the validation of task level capabilities, including point reaching and joint rotation. The section concludes with a discussion of variable stiffness demonstrations and results, which showcases the robot's adaptability in real-world applications.

\subsection{Experiment setup}

The experiments described in this section were conducted by two experimental setups, as illustrated in Figure \ref{fig:unlock_stiff_experiment_setup}A and \ref{fig:lock_stiffness_experiment}A. This setup includes the PLC robot system and its actuation configuration, the controller system, external force sensors, a two-axis motor system, and a motion capture system.

The PLC robot used in the experiments is composed of five modules, manufactured by ABS resin. Each module is independently driven by a linear actuator to enable stiffness adjustment (Fig. \ref{fig:lock_stiffness_experiment}A). The robot’s actuation system is comprised of five linear actuators and a rotary actuator, all of which are arranged beneath the robot to control the cable-driven actuation and joint rotation.

Additionally, a force sensor and a motion capture system are employed for data acquisition during the experiments. The force sensor is mounted on the end effector of the two-axis motor system, enabling precise measurement of force along the parallel surface. The two-axis motor system also allows for accurate displacement control in predetermined directions, providing reliable ground truth data for both force and displacement through open-loop movements. The motion capture system (OptiTracker) is used to track the spatial movement of reflective markers attached to the robot. This system is primarily utilized in trajectory and target-reaching tasks to capture the position of the robot’s end-effector. Accurate displacement and force data are collected based on the motor encoder and 1-dof force sensor, as shown in the Fig. \ref{fig:unlock_stiff_experiment_setup}A. 

\subsection{Stiffness Model Validation}

As mentioned in Section \ref{sec:stiffness modeling}, hardware experiments are conducted to validate the stiffness model. Experiments on a single PLC robot segment targets to verify the accuracy of the theoretical firmed stiffness via Section \ref{subsec:locking stiffness} and loosening threshold via Section \ref{subsec:unlocked stiffness}. They will also determine if the anticipated loosening stiffness in section \ref{subsec:unlocked stiffness} caused by tendon stretch presents. In addition, multi-segment stiffness test is also carried out to examine the overall behavior of the full PLC robot body.

\subsubsection{Single-Segment Stiffness}

\begin{figure}[!t] 
    \centering
    \includegraphics[width=\linewidth]{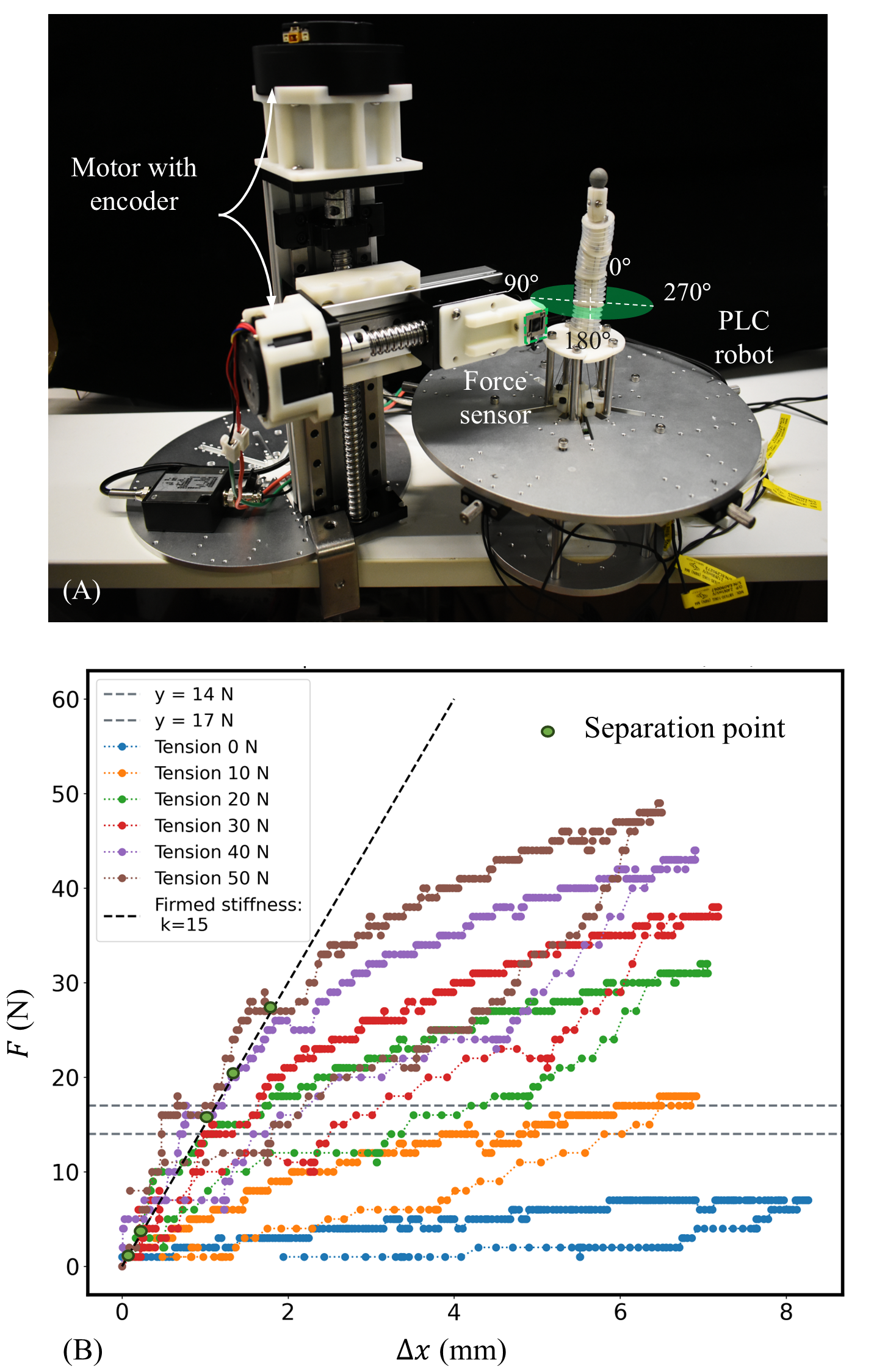}
    \caption{(A) The experimental setup for collecting single segment force-deformation data. (B) The force–deformation plot of a single segment. The firmed stiffness is represented by a dashed line. As the tension increases, the PLC unit exhibits greater stiffness, resulting in a higher separation point under increasing pushing force.}
    \label{fig:unlock_stiff_experiment_setup}
\end{figure}

Experiments in this part are conducted only on the lowest segment of the PLC robot and leave tendons on other segments loosening (Fig. \ref{fig:unlock_stiff_experiment_setup}A). We collect the displacement and applied force data of the lowest segment under various tendon tension, such that attaining the relation between the tension, displacement $\Delta x$, and force $\Delta F$. Notice the locking ring between the lowest and the white platform holding the PLC robot is the one detached during loosening stiffness. The segment is first pushed to a horizontal deformation between $6$ and $8$ mm, we then slowly retrieve the pushing motor to allow the segment to return to its neutral position.

Using the parameters in Table \ref{table: major parameters}, the firmed stiffness is predicted to be $14$ to $17$ N/mm. This predicted value is a range because of the segment's anisotropy and uncertainty in the material's Young's modulus. Tendon connection points are distributed on a 6mm-radius circle on the lower ring . The pushing point $\mathbf{p}_{ext}$ approximately is 20 mm right above the fulcrum $\mathbf{p}_{fu}$ and the pushing force $\mathbf{F}_{ext}$ lies on the x-y plane. Based on these facts, the loosening threshold on the $\mathbf{F}_{ext}$ magnitude can be estimated as $F_{th}\approx\frac{3}{5}|\mathbf{T}|$.

Fig. \ref{fig:unlock_stiff_experiment_setup}B shows the displacement-force curves. Hysteresis presented due to internal friction of the mechanical structure, including tendons and variable stiffness mechanism resulted material deformation, and elastic and inelastic interactions between the tendon and the robot. The dash line is the firmed stiffness of the single segment, where curves under different tensions $\mathbf{T}$ follows when the deformation $\Delta x$ is small. It is measured to be around 15 N/mm, which matches the theoretical prediction quite well. As the external force $\Delta F$ continues to grow, separation points appear where the curves leave the dash line and start to follow a lower growing ratio. This is the loosening threshold. Via visual observation of Fig. \ref{fig:unlock_stiff_experiment_setup}B, the loosening thresholds for tendon tensions $30$, $40$, $50$ N are respectively in intervals $[14, 19]$ N, $[24, 26]$ N, $[28, 33]$ N. This also agrees with the theoretical estimation $F_{th}=\frac{3}{5}|\mathbf{T}|$.

All curves, except for the one with tension $|\mathbf{T}|=0$, are approximately parallel to each other after the separation point. This behavior occurs before entering the hysteresis phase and is consistent with the tendon's linear strain response. This is expected via the model in Section \ref{subsec:unlocked stiffness} where tension $|\mathbf{T}|$ shall not affect the tendon's elasticity before it exceeds the tendon's endurance. When $|\mathbf{T}|=0$, the tendon connecting the lowest ring is also loose, so the overall elasticity is based only on the flexible shaft at the center of the PLC robot. This explains the extraordinary flatness of the $|\mathbf{T}|=0$ curve. 

The validation of twisting stiffness is shown in Fig.\ref{fig:lock_stiffness_experiment}C. Both the resin spine and the origami skin exhibit better linearity near zero torque. During the experiment, the origami skin begins to buckle when the applied torque approaches 0.5Nm, which is also supported in the data.

\subsubsection{Multi-Segment Stiffness}

Experiments in this part focus on validating the combined firmed stiffness on multi-segment PLC robot end-effector under various configurations.
 
\begin{figure}[!t] 
    \centering
    \includegraphics[width=\linewidth]{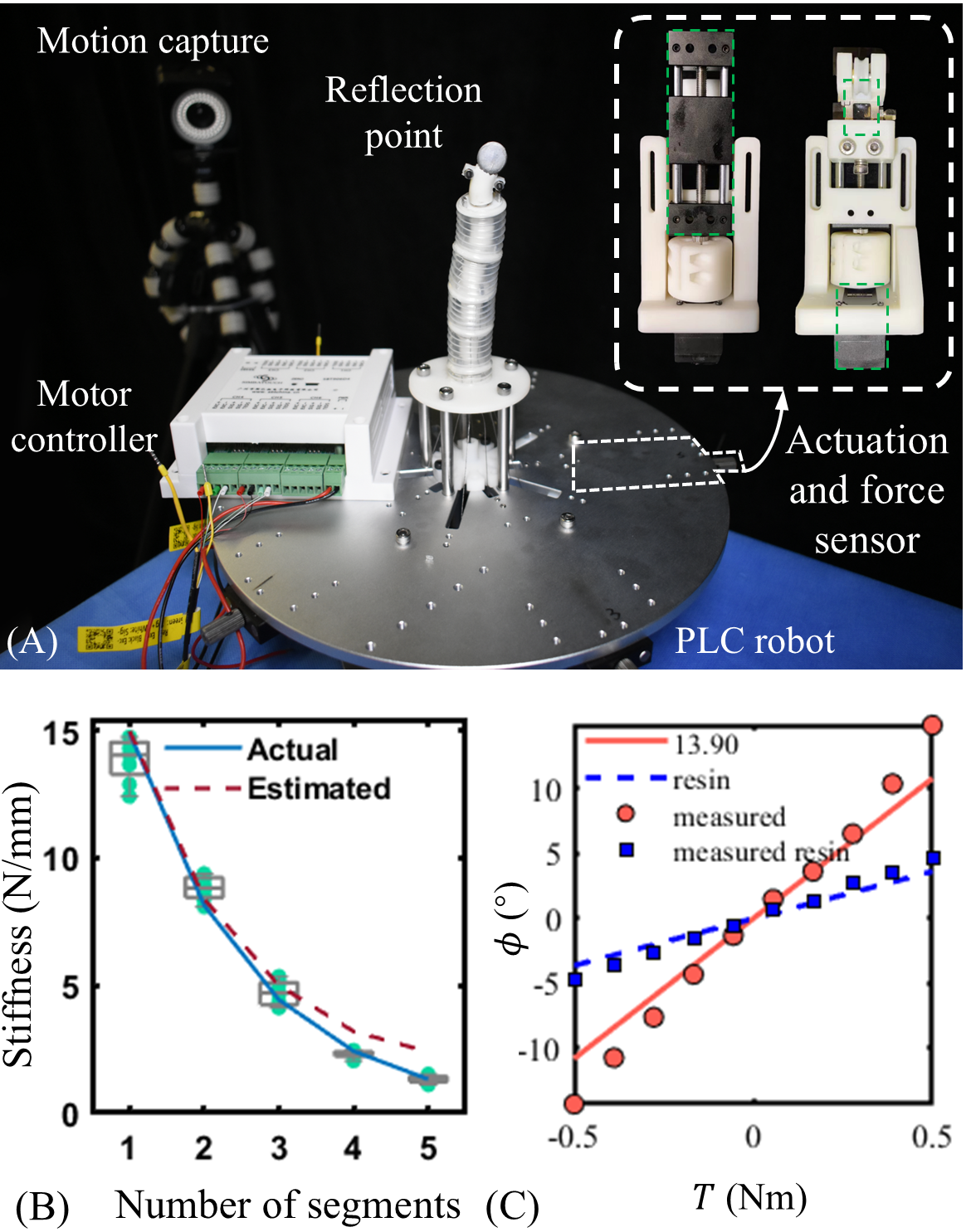}
    \caption{ Firmed stiffness results. (A) The experimental setup for firmed stiffness validation and task space planning. (B) Results of the firmed stiffness verification with different segments. The tension of each segment is 50 N. (C) The twist experiment on the resin spine and origami skin. The origami skin happens slight buckling when the torque increases.}
    \label{fig:lock_stiffness_experiment}
\end{figure}

First, we examine the overall stiffness across different numbers of segments. We randomly selected a specific configuration of the PLC robot and measured the stiffness on first to the fifth segment's top bearing. For each segment, stiffness was obtained in eight directions within the plane perpendicular to \(v_i\), corresponding to angles \(q_i = [0, \pi/4, \pi/2, \dots, 7\pi/4]\). As illustrated in Figure \ref{fig:lock_stiffness_experiment}B, the x-axis represents the segment number \(n\), while the y-axis represents the stiffness. The stiffness measurements for each segment in the eight directions are plotted as individual data points. The average values (indicated by the middle line of the box plot) were fitted with a function to estimate a numerical stiffness profile.

In addition, theoretical estimates of the average stiffness, based on the same robot configuration and measurement angles, were similarly fitted and plotted. It was observed that the theoretical estimates tend to be slightly higher than the measured values, with the discrepancy increasing as the number of segments increases. This difference can be attributed to the presence of backlash in the robot, which accumulates with more segments due to manufacturing tolerances, affecting the end stiffness. This phenomenon is akin to the gear backlash observed in mechanical reducers.

As the PLC robot is composed of multiple components, the complexity affects some modeling parameters, such as Young's modulus. Therefore, physical parameters used in modeling process are given with a range to compensate this error, which could lead to larger variation when the number of segments increases (Fig. \ref{fig:lock_stiffness_experiment}B and \ref{fig:lock_stiffness_experiment}C).

The end point stiffness under different configurations focused on the variation of stiffness in eight directions at the end of the robot, considering both four-segment and five-segment PLC robot configurations. Five random configurations were selected, and stiffness was measured at the robot's end in the eight directions perpendicular to \(v_i\) at angles \(q_i = [0, \pi/4, \pi/2, \dots, 7\pi/4]\). Figures \ref{fig:lock_stiffness_experiment}C shows the results, where the x-axis represents the eight directions (Fig. \ref{fig:lock_stiffness_experiment}A) and the y-axis represents the stiffness. The measured data and theoretical simulation data for the five different configurations are plotted together.

Two key observations can be made from these results: first, the theoretical stiffness is consistently higher than the actual measured stiffness; second, for each configuration, similar trends in firmed stiffness variation can be observed between the theoretical and actual data.

\subsection{Task Space Planning}

This section presents experiments demonstrating the fundamental capabilities of the PLC robot within the task space. These capabilities include point reaching task and continuous rotation of the PLC robot within the task space.

\begin{figure}[!t] 
    \centering
    \includegraphics[width=\linewidth]{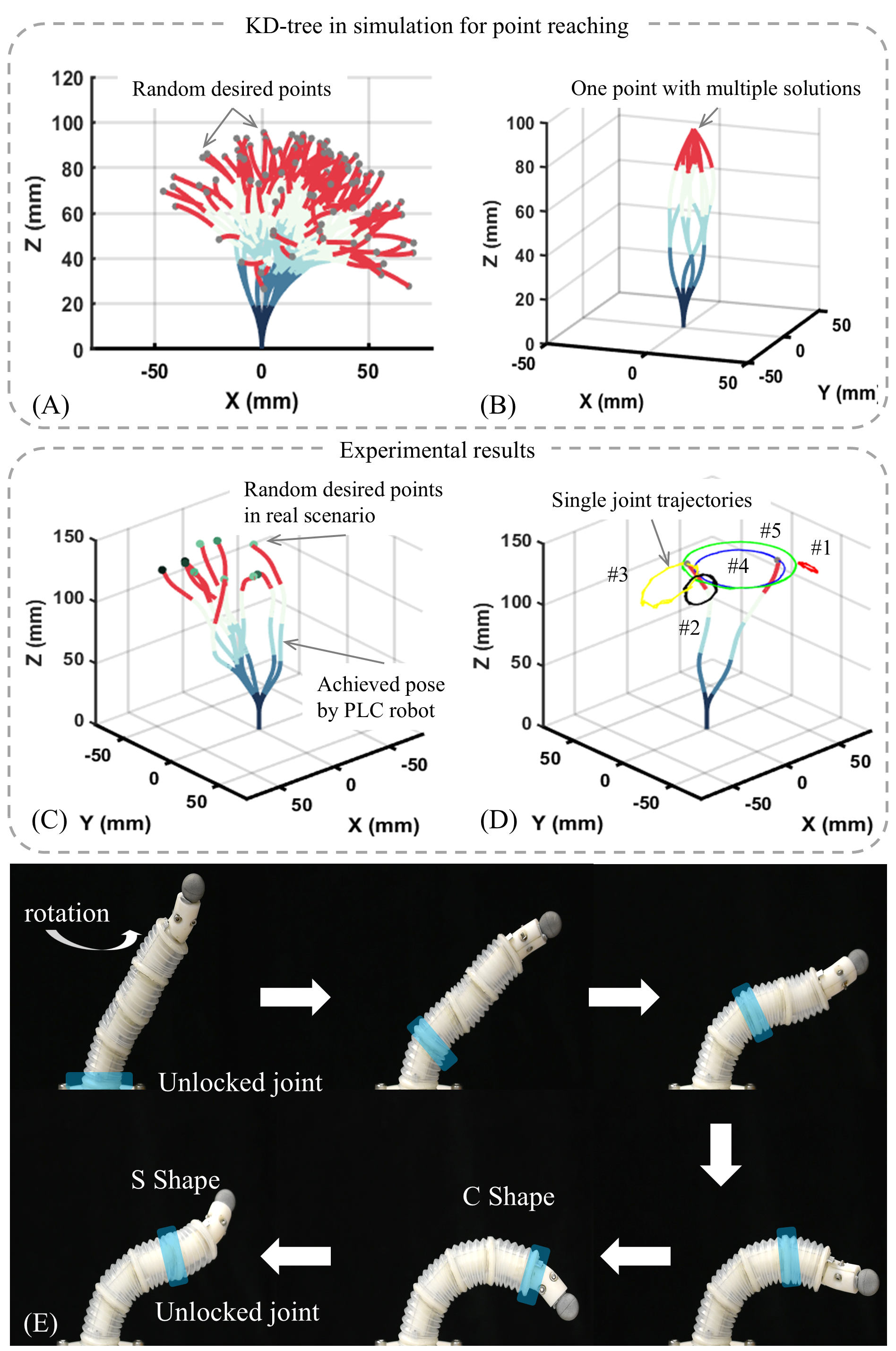}
    \caption{Planning of the PLC robot. (A) The theoretical KD-tree planning results in task space. (B) Illustration of multiple solution condition in the inverse kinematics. (C) Point reaching results in real scenario. (D) Rotation trajectory by independently moving five joints in five different configurations. (E) Demonstration of achieving C-shape and S-shape from a random joint configuration. }
    \label{fig:point_reach}
\end{figure}

\subsubsection{Point Reaching based on KD-Tree Planning}

The point reaching task involves guiding the continuum robot's end effector to a specified target point in space. Unlike other continuum robots, the PLC robot's workspace is discrete, meaning significant stiffness changes can only be achieved at these discrete spatial points. However, as discussed in the methods section, while these operational points are discrete, they become densely packed within the workspace as the number of modules and teeth increases, e.g. Fig. \ref{fig:kinematics}C.

First, we theoretically verify the point reaching problem in simulation, as shown in the Fig. \ref{fig:point_reach}A and \ref{fig:point_reach}B. The target points are randomly selected within the workspace of the 5-segmented PLC robot, which is represented as grey points in the Fig. \ref{fig:point_reach}A. Applying the KD-tree planning algorithm (Algorithm \ref{algo:inverse}) and remove residual configuration solutions, the PLC robot go to a specific configuration that make its end effector as close to the target point as possible. Besides, there are some multi-solutions conditions as shown in the Fig. \ref{fig:point_reach}B. When assigning a target point, the algorithm pick up the closest point to the target point, however, there are more than one solution to that closest point. In planning, we randomly remove the residual solutions.

The point reach experimental results are shown in the Fig. \ref{fig:point_reach}C. The robot configurations in this figure are collected from the real robot. By assigning the robot with a target point, the planner gives a desired robot configuration and controls the PLC robot move to the desired configuration. Afterwards, the actual configuration of the robot are feedback to present in the figure. The target points with different colors are used to avoid ambiguity, as some points are close. The theoretical point reach accuracy can be expressed as
\begin{equation}
    \max_i\{\min_{j}\{\text{d}\left(p_i,p_j\right)\}\}, p_i, p_j\in \mathcal{P}
\end{equation}
where $\mathcal{P}$ is reachable point set, $\text{d}\left(\cdot\right)$ is the Euclidean distance between two points.

\subsubsection{Independent Joint Rotation}

The independent joint rotation experiments show the continuous movement capability of the PLC robot. We record the rotation under five different configurations (Fig. \ref{fig:point_reach}D), where \#1 represents the rotation of the top most module, \#2{-\#5 represent the modules from top to down. In the continuous rotation, the end effector draw a circle trajectory, although it can only stay in discrete positions on this trajectory to varying stiffness. With this decoupled independent motion of each joint, we can achieve any configuration and reach any point in the workspace in programmable motion, as shown in the Fig. \ref{fig:point_reach}E.

\subsection{Functional Demonstrations in Constrained Environments and Adaptive Grasping}

To validate the practical utility of the PLC system, we conducted a series of functional demonstrations that span adaptive grasping, compliant manipulation, and task execution in constrained environments, as shown in Fig. 15. A two-finger PLC gripper, with each finger composed of three serially connected modules, was used to grasp objects under variable stiffness conditions. In the low-stiffness mode (Fig. 15A), the gripper conformed to deformable and irregular objects such as sponges and toys, leveraging compliance to maximize contact area. When switched to high-stiffness mode (Fig. 15B), the same gripper securely lifted heavy items such as a 2 kg water bottle and a 1 kg weight, demonstrating its ability to transition between gentle and firm grasping strategies.

Fig. 15C highlights the fine manipulation capabilities enabled by discrete stiffness control at the terminal PLC module. By locking or releasing only the last unit, both fingertip and whole-finger rotations were achieved, enabling compliant fine in-hand reconfiguration suitable for tool use or dexterous interaction. In addition, a sixteen-unit PLC manipulator was deployed for a complex task involving pipe insertion and distal screwing (Fig. 15D). The system maintained compliance during insertion to conform to the pipe geometry, then selectively stiffened the front modules to deliver sufficient torque for the screwing operation. This scenario demonstrates the system’s capacity for spatially distributed stiffness control and structural support during high-force manipulation in confined environments.

Together, these demonstrations confirm that the PLC system provides programmable, segment-level stiffness modulation to support a wide range of robotic tasks requiring both compliance and rigidity. The results further validate the scalability and reconfigurability of the PLC architecture in both gripper and manipulator formats.

\begin{figure}[!t] 
    \centering
    \includegraphics[width=\linewidth]{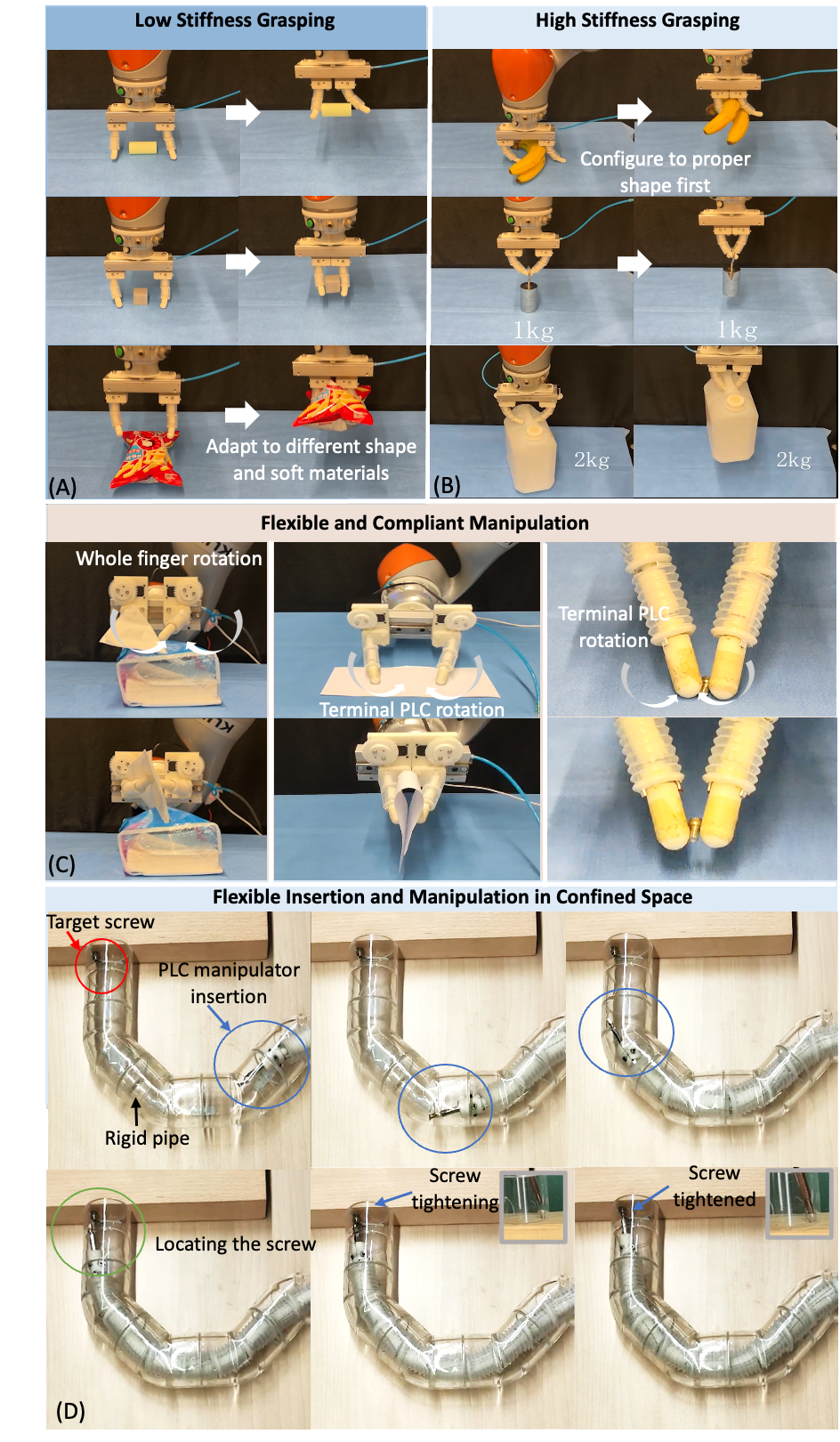}
    \caption{\
(A) A two-finger PLC gripper performing adaptive low stiffness grasping.
(B) A two-finger PLC gripper demonstrating robust high stiffness grasping.
(C)  Whole-finger and terminal PLC actuation for compliant and fine in-hand manipulation.
(D) A sixteen-unit PLC manipulator performing pipe insertion and screw tightening.}
    \label{fig:demo}
\end{figure}

\subsection{Limitation Discussion}

Despite the demonstrated capabilities of the PLC system, several limitations remain. First, the current stiffness control is realized through a continuous tendon routed across multiple units. This introduces mechanical coupling between segments, which can lead to interference during stiffness transitions and reduce the system’s control resolution and modularity. Second, the dynamic characteristics of stiffness modulation—such as switching latency, repeatability under cyclic loading, and closed-loop response—have not yet been fully characterized. Third, structural tolerances and mechanical backlash accumulate along long PLC chains, degrading end-effector accuracy and global stiffness consistency. These issues will be addressed in future discrete PLC units with enhanced designs, in which each unit incorporates embedded actuators for independent stiffness tuning, enabling discrete stiffness modulation, improving modular unit performance characterization, and minimizing cross-unit error accumulation.

\section{Conclusion and Future Work}

This paper presents a structure-centric robotic framework based on Programmable Locking Cells (PLCs), enabling high-range discrete stiffness modulation and programmable morphology through tendon-actuated mechanical engagement. Each PLC unit can continuously transition between compliant and rigid states with up to 950\% stiffness variation, and multiple units can be assembled into scalable robotic systems with spatially distributed stiffness control.

We developed analytical models capturing both firmed and loosening stiffness states, characterized their anisotropic behavior, and validated these models through extensive single- and multi-segment experiments. A discrete motion planning strategy based on k-d tree search was introduced to address inverse kinematics in non-continuous configuration spaces. Two functional prototypes—an adaptive PLC gripper and a pipe-traversing manipulator—demonstrated the system’s capabilities in compliant grasping, in-hand manipulation, and high-force tasks in constrained environments. These results collectively validate the PLC architecture as a robust and versatile foundation for variable-stiffness robotic systems.

In future work, we aim to transition from a centralized tendon-routing design to a fully decentralized PLC architecture, with each unit equipped with embedded actuation for independent stiffness control. This approach will eliminate inter-segment dependencies, enhance modularity, and enable true plug-and-play reconfiguration. Efforts will also focus on miniaturizing the actuation module to improve portability and facilitate seamless integration in multi-unit PLC assemblies. Furthermore, we plan to systematically characterize the system’s dynamic behavior, including cyclic fatigue resistance, responsiveness in dynamic manipulation tasks, and hysteresis effects, to inform long-term reliability improvements and control optimization. These advancements will support the deployment of PLC-based robots in applications requiring adaptive compliance, robustness, and scalable reconfigurability, such as surgical intervention, long-reach exploration, and human-robot collaboration.

\appendices

\section{Workspace Omnivariance}
\label{appenA}

The omnivariance is attained from the covariance tensor. The covariance matrix, $\Sigma$, is computed from the point cloud's 3D coordinates, capturing how points are distributed relative to one another along each axis. For a given set of 3D points $(x_i, y_i, z_i)$, the covariance matrix is a 3×3 matrix, representing the variance and covariance of points in the x, y, and z dimensions.
\begin{equation}
    \Sigma=\frac{1}{N} \sum_{i \in \mathcal{P}}\left(\mathbf{p}_i-\overline{\mathbf{p}}\right)\left(\mathbf{p}_i-\overline{\mathbf{p}}\right)^{\top}
\end{equation}
where $\mathbf{p}_i=(x_i,y_i,z_i)^T$ is a point in the point cloud. $\mathcal{P}$ denotes the cloud set. $\overline{\mathbf{p}}$ is the mean of all points in the cloud. $N$ is the number of points. Omnivariance of the working space cloud is attained with
\begin{equation}
    \Theta = \left(\lambda_1\cdot\lambda_2\cdot\lambda_3\right)^{\frac{1}{3}}
\end{equation}
where the $\lambda_1 \geq \lambda_2 \geq \lambda_3 \geq 0$ is the eigenvalues of the covariance matrix $\Sigma$. 

\begin{figure}[t] 
    \centering
    \includegraphics[width=\linewidth]{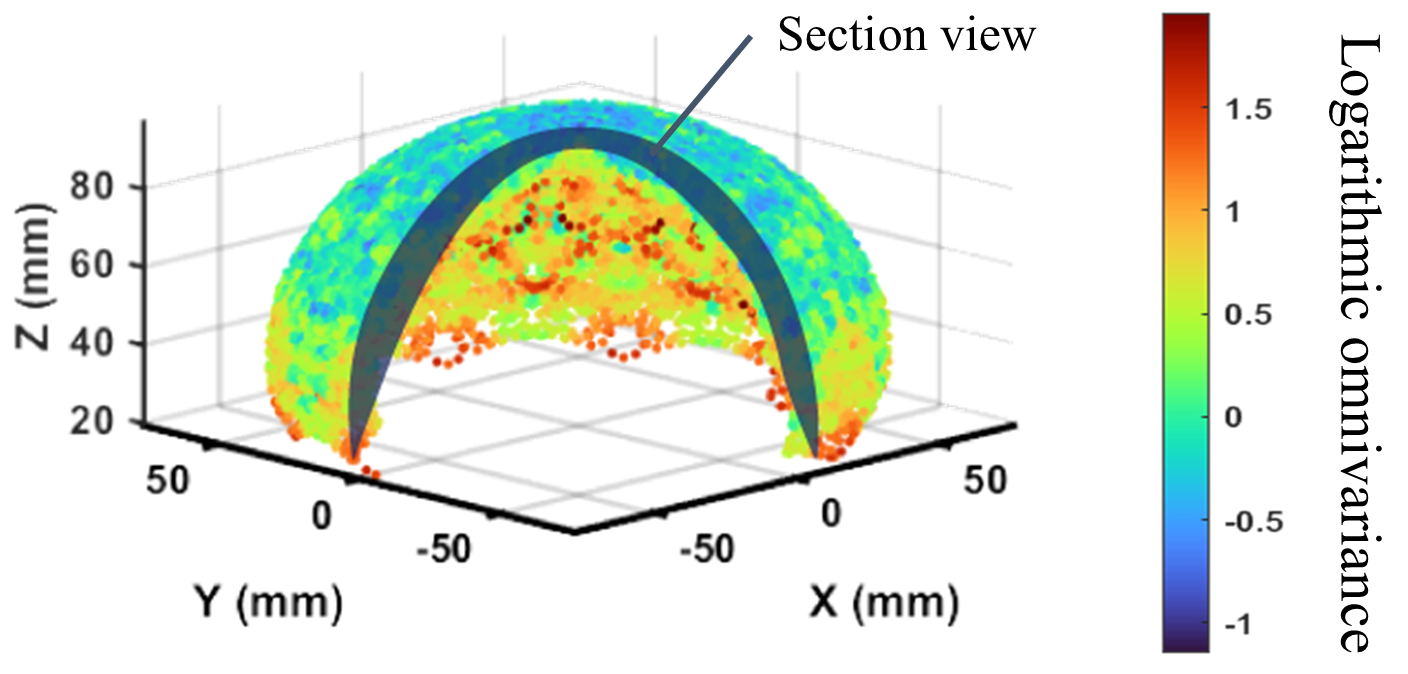}
    \caption{The logarithmic omnivariance over the working space point cloud.}
    \label{fig:omnivariance}
\end{figure}

\section{Normalized Stiffness}
\label{apdx:notmalized stiffness}

Beside mechanism designs, stiffness of a continuum robot is also determined by its length and cross section area. As such, a method for normalizing stiffness using the effective radius $R$ and length $L$ is proposed for comparing the maximal and minimal stiffness across different robot designs. Our primary focus, which is the bending payload of the continuum robot, is similar to the objective of the cantilever beam model. Based on the Euler-Bernoulli beam theory \cite{beer1992mechanics}, the bending stiffness of cantilever beam is

\begin{equation}
    k = \frac{3EI}{L^3},\quad I = \frac{\pi R^4}{64}
\label{eq:beam_stiffness}
\end{equation}
where $L$ is the length of the beam and $I$ is denoted as the second area momentum of a cylinder. Inspired by the fact that the Young's modulus $E$ is the sole component independent of the beam size in (\ref{eq:beam_stiffness}), the following stiffness normalization method is proposed given the original stiffness $K_b$. It is important to note that the Euler-Bernoulli beam theory yields normalized stiffness values that are independent of radius and length only in the case of cylindrical robots with uniformly distributed material properties. Consequently, such normalization serves primarily for visualization purposes, and some degree of bias is inherent in the context of variable-stiffness robot designs. In this work, we focus on comparing the ratio of maximal to minimal stiffness, which, according to Eq. (\ref{eq:beam_stiffness}), remains unaffected by the normalization process. A detailed analysis of the accuracy of the Euler-Bernoulli beam theory is beyond the scope of this paper.

\begin{equation}
    K_{nb} = K_b\frac{ L^3}{R^4} 
\label{eq:normalised_bending_stiffness} 
\end{equation}

The original and normalized stiffness, both in the maximal and minimal configurations, of recent varying stiffness robot literature are summarized in Table \ref{tab:stiff-compare}. {Fig. \ref{fig:summarize scatter} is a zoomed-in view of the minimal-to-maximal stiffness region, which is closer to our proposed design. Data points outside this region are not visible in Fig. \ref{fig:summarize scatter} and are indicated with an asterisk (*) in Table \ref{tab:stiff-compare}. The original stiffness has unit $N/mm$ and the normalized stiffness has unit $N/mm^2$. Normalized stiffness may not be available if the robot length or radius is not reported in the literature.

\begin{table}[t]
\centering
\caption{Summary of Recent Varying Stiffness Robots}
\begin{tabular}{p{24mm} | p{20mm} | p{20mm} | p{5mm}}
    Reference & Max Stiffness\hfill\break Original / \hfill\break Normalized & Min Stiffness\hfill\break Original / \hfill\break Normalized & Ratio \\
    \hline\hline
    \textbf{Ours} & \textbf{8.07} / \textbf{252.19} & \textbf{0.85} / \textbf{26.56} & \textbf{9.5} \\
    Chen et al* \cite{chen2021tele}              & 1.73 / 0.0072      & 0.75 / 0.0031             & 2.3 \\
    Zhou et al \cite{zhou2021bio}                & 1 / 35.12          & 0.40 / 14.05              & 2.5 \\
    Farooq \& Ko \cite{farooq2018stiffness}      & 0.064 / 1.02e2     & 0.016 / 25.60             & 4.0 \\
    Brancadoro et al \cite{brancadoro2019toward} & 0.057 / 2.58       & 0.022 / 1.00              & 2.6 \\
    Jiang et al \cite{jiang2012design}           & 0.29 / 29.70       & 0.07 / 7.17               & 4.1 \\
    Kim et al \cite{kim2013stiffness}            & 0.48 / 85.25       & 0.16 / 28.53              & 3.0 \\
    Li et al* \cite{li2019flexible}              & 3.5 / NA           & 0.50 / NA                 & 7.0 \\
    Li et al* \cite{li2020single}                & 0.88 / 5.37e3      & 0.40 / 2.43e3             & 2.2 \\
    Liu et al \cite{liu2021positive}             & 2 / 8.75           & 0.55 / 2.41               & 3.6 \\
    Sadati et al* \cite{sadati2015stiffness}     & 0.073 / 0.024      & 0.04 / 0.13               & 1.8 \\
    Sadati et al* \cite{sadati2018three}         & 0.039 / 0.031      & 0.0056 / 0.0045           & 7.0 \\
    Stilli et al \cite{stilli2014shrinkable}     & 0.63 / 2.90e2      & 0.32 / 1.46e2             & 2.0 \\
    Yang et al \cite{yang2020geometric}          & 0.033 / 1.73e2     & 0.013 / 67.98             & 2.5 \\
    Sun et al* \cite{sun2020hybrid}              & 0.042 / 3.01e3     & 0.025 / 1.79e3            & 1.7 \\
    Moses et al \cite{moses2013continuum}        & 0.23 / 22.75       & 0.17 / 16.9               & 1.3 \\
    Zhao et al* \cite{zhao2019soft}              & 0.47 / NA          & 0.14 / NA                 & 3.4 \\
    Zhu and Hu* \cite{zhu2021controllable}       & 0.42 / NA          & 0.067 / NA                & 6.3 \\
    Kim et al \cite{kim2013novel}                & 0.21 / 8.96e2      & 0.11 / 4.59e2             & 2.0
\end{tabular}
\label{tab:stiff-compare}
\end{table}



\bibliographystyle{IEEEtran}
\bibliography{main}

\begin{thebibliography}{10}
\providecommand{\url}[1]{#1}
\csname url@samestyle\endcsname
\providecommand{\newblock}{\relax}
\providecommand{\bibinfo}[2]{#2}
\providecommand{\BIBentrySTDinterwordspacing}{\spaceskip=0pt\relax}
\providecommand{\BIBentryALTinterwordstretchfactor}{4}
\providecommand{\BIBentryALTinterwordspacing}{\spaceskip=\fontdimen2\font plus
\BIBentryALTinterwordstretchfactor\fontdimen3\font minus
  \fontdimen4\font\relax}
\providecommand{\BIBforeignlanguage}[2]{{%
\expandafter\ifx\csname l@#1\endcsname\relax
\typeout{** WARNING: IEEEtran.bst: No hyphenation pattern has been}%
\typeout{** loaded for the language `#1'. Using the pattern for}%
\typeout{** the default language instead.}%
\else
\language=\csname l@#1\endcsname
\fi
#2}}
\providecommand{\BIBdecl}{\relax}
\BIBdecl

\bibitem{kim2013mri}
Y.-J. Kim, S.~Cheng, S.~Kim, and K.~Iagnemma, ``A stiffness-adjustable
  hyperredundant manipulator using a variable neutral-line mechanism for
  minimally invasive surgery,'' \emph{IEEE Transactions on Robotics}, vol.~30,
  no.~2, pp. 382--395, 2013.

\bibitem{shan2023variable}
Y.~Shan, Y.~Zhao, H.~Wang, L.~Dong, C.~Pei, Z.~Jin, ..., and T.~Liu, ``Variable
  stiffness soft robotic gripper: Design, development, and prospects,''
  \emph{Bioinspiration \& Biomimetics}, vol.~19, no.~1, p. 011001, 2023.

\bibitem{manti2016review}
M.~Manti, V.~Cacucciolo, and M.~Cianchetti, ``Stiffening in soft robotics: A
  review of the state of the art,'' \emph{IEEE Robotics \& Automation
  Magazine}, vol.~23, no.~3, pp. 93--106, 2016.

\bibitem{polygerinos2017soft}
P.~Polygerinos, Z.~Wang, K.~C. Galloway, R.~J. Wood, and C.~J. Walsh, ``Soft
  robotic glove for combined assistance and at-home rehabilitation,''
  \emph{Robotics and Autonomous Systems}, vol.~73, pp. 135--143, 2015.

\bibitem{jiang2012granular}
A.~Jiang, G.~Xynogalas, P.~Dasgupta, K.~Althoefer, and T.~Nanayakkara, ``Design
  of a variable stiffness flexible manipulator with composite granular jamming
  and membrane coupling,'' in \emph{IEEE/RSJ International Conference on
  Intelligent Robots and Systems (IROS)}, 2012, pp. 2922--2927.

\bibitem{wolf2015review}
S.~Wolf, G.~Grioli, O.~Eiberger, W.~Friedl, M.~Grebenstein, H.~Hoppner,
  E.~Burdet, D.~G. Caldwell, R.~Carloni, M.~G. Catalano, and D.~Lefeber,
  ``Variable stiffness actuators: Review on design and components,''
  \emph{IEEE/ASME Transactions on Mechatronics}, vol.~21, no.~5, pp.
  2418--2430, 2015.

\bibitem{dou2021soft}
W.~Dou, G.~Zhong, J.~Cao, Z.~Shi, B.~Peng, and L.~Jiang, ``Soft robotic
  manipulators: Designs, actuation, stiffness tuning, and sensing,''
  \emph{Advanced Materials Technologies}, vol.~6, no.~9, p. 2100018, 2021.

\bibitem{yim2000polybot}
M.~Yim, D.~G. Duff, and K.~D. Roufas, ``Polybot: a modular reconfigurable
  robot,'' in \emph{Proceedings 2000 ICRA. Millennium Conference. IEEE
  International Conference on Robotics and Automation}, vol.~1.\hskip 1em plus
  0.5em minus 0.4em\relax IEEE, 2000, pp. 514--520.

\bibitem{yim2007modular}
M.~Yim, W.~M. Shen, B.~Salemi, D.~Rus, M.~Moll, H.~Lipson, and G.~S.
  Chirikjian, ``Modular self-reconfigurable robot systems [grand challenges of
  robotics],'' \emph{IEEE Robotics \& Automation Magazine}, vol.~14, no.~1, pp.
  43--52, 2007.

\bibitem{seo2019modular}
J.~Seo, J.~Paik, and M.~Yim, ``Modular reconfigurable robotics,'' \emph{Annual
  Review of Control, Robotics, and Autonomous Systems}, vol.~2, no.~1, pp.
  63--88, 2019.

\bibitem{rao2021model}
P.~Rao, Q.~Peyron, S.~Lilge, and J.~Burgner-Kahrs, ``How to model tendon-driven
  continuum robots and benchmark modelling performance,'' \emph{Frontiers in
  Robotics and AI}, vol.~7, p. 630245, 2021.

\bibitem{toshimitsu2021sopra}
Y.~Toshimitsu, K.~W. Wong, T.~Buchner, and R.~Katzschmann, ``Sopra: Fabrication
  \& dynamical modeling of a scalable soft continuum robotic arm with
  integrated proprioceptive sensing,'' in \emph{2021 IEEE/RSJ International
  Conference on Intelligent Robots and Systems (IROS)}.\hskip 1em plus 0.5em
  minus 0.4em\relax IEEE, 2021, pp. 653--660.

\bibitem{shiva2016tendon}
A.~Shiva, A.~Stilli, Y.~Noh, A.~Faragasso, I.~De~Falco, G.~Gerboni,
  M.~Cianchetti, A.~Menciassi, K.~Althoefer, and H.~A. Wurdemann,
  ``Tendon-based stiffening for a pneumatically actuated soft manipulator,''
  \emph{IEEE Robotics and Automation Letters}, vol.~1, no.~2, pp. 632--637,
  2016.

\bibitem{kim2017active}
Y.~Kim, S.~S. Cheng, and J.~P. Desai, ``Active stiffness tuning of a
  spring-based continuum robot for mri-guided neurosurgery,'' \emph{IEEE
  Transactions on Robotics}, vol.~34, no.~1, pp. 18--28, 2017.

\bibitem{an2023active}
X.~An, Y.~Cui, H.~Sun, Q.~Shao, and H.~Zhao, ``Active-cooling-in-the-loop
  controller design and implementation for an sma-driven soft robotic
  tentacle,'' \emph{IEEE transactions on robotics}, vol.~39, no.~3, pp.
  2325--2341, 2023.

\bibitem{lin2022modular}
B.~Lin, J.~Wang, S.~Song, B.~Li, and M.~Q.-H. Meng, ``A modular lockable
  mechanism for tendon-driven robots: design, modeling and characterization,''
  \emph{IEEE Robotics and Automation Letters}, vol.~7, no.~2, pp. 2023--2030,
  2022.

\bibitem{kim2019continuously}
J.~Kim, W.-Y. Choi, S.~Kang, C.~Kim, and K.-J. Cho, ``Continuously variable
  stiffness mechanism using nonuniform patterns on coaxial tubes for continuum
  microsurgical robot,'' \emph{IEEE Transactions on Robotics}, vol.~35, no.~6,
  pp. 1475--1487, 2019.

\bibitem{li2015ultimate}
Y.~Li and Y.~Chen, ``The ultimate hyper redundant robotic arm based on
  omnidirectional joints,'' in \emph{2015 IEEE International Conference on
  Mechatronics and Automation (ICMA)}.\hskip 1em plus 0.5em minus 0.4em\relax
  IEEE, 2015, pp. 1840--1845.

\bibitem{jitosho2023passive}
R.~Jitosho, S.~Sim{\'o}n-Trench, A.~M. Okamura, and B.~H. Do, ``Passive shape
  locking for multi-bend growing inflated beam robots,'' in \emph{2023 IEEE
  International Conference on Soft Robotics (RoboSoft)}, 2023, pp. 1--6.

\bibitem{kim2017novel}
J.~Kim, C.~Kim, S.~Kang, and K.~Cho, ``A novel variable stiffness mechanism for
  minimally invasive surgery using concentric anisotropic tube structure,'' in
  \emph{The Hamlyn Symposium on Medical Robotics}, 2017, p.~43.

\bibitem{kim2013novel}
Y.-J. Kim, S.~Cheng, S.~Kim, and K.~Iagnemma, ``A novel layer jamming mechanism
  with tunable stiffness capability for minimally invasive surgery,''
  \emph{IEEE Transactions on Robotics}, vol.~29, no.~4, pp. 1031--1042, 2013.

\bibitem{hassan2017active}
T.~Hassan, M.~Cianchetti, B.~Mazzolai, C.~Laschi, and P.~Dario, ``Active-braid,
  a bioinspired continuum manipulator,'' \emph{IEEE Robotics and Automation
  Letters}, vol.~2, no.~4, pp. 2104--2110, 2017.

\bibitem{fan2022novel}
Y.~Fan, D.~Liu, and L.~Ye, ``A novel continuum robot with stiffness variation
  capability using layer jamming: Design, modeling, and validation,''
  \emph{IEEE Access}, vol.~10, pp. 130\,253--130\,263, 2022.

\bibitem{kim2013stiffness}
Y.-J. Kim, S.~Cheng, S.~Kim, and K.~Iagnemma, ``A stiffness-adjustable
  hyperredundant manipulator using a variable neutral-line mechanism for
  minimally invasive surgery,'' \emph{IEEE transactions on robotics}, vol.~30,
  no.~2, pp. 382--395, 2013.

\bibitem{al2017design}
L.~A. Al~Abeach, S.~Nefti-Meziani, and S.~Davis, ``Design of a variable
  stiffness soft dexterous gripper,'' \emph{Soft robotics}, vol.~4, no.~3, pp.
  274--284, 2017.

\bibitem{stilli2014shrinkable}
A.~Stilli, H.~A. Wurdemann, and K.~Althoefer, ``Shrinkable,
  stiffness-controllable soft manipulator based on a bio-inspired antagonistic
  actuation principle,'' in \emph{2014 IEEE/RSJ International Conference on
  Intelligent Robots and Systems}.\hskip 1em plus 0.5em minus 0.4em\relax IEEE,
  2014, pp. 2476--2481.

\bibitem{choi2020design}
J.~Choi, S.~H. Ahn, C.~Kim, J.-H. Park, H.-Y. Song, and K.-J. Cho, ``Design of
  continuum robot with variable stiffness for gastrointestinal stenting using
  conformability factor,'' \emph{IEEE Transactions on Medical Robotics and
  Bionics}, vol.~2, no.~4, pp. 529--532, 2020.

\bibitem{yang20163d}
Y.~Yang, Y.~Chen, Y.~Wei, and Y.~Li, ``3d printing of shape memory polymer for
  functional part fabrication,'' \emph{The International Journal of Advanced
  Manufacturing Technology}, vol.~84, pp. 2079--2095, 2016.

\bibitem{wei2021analysis}
X.~Wei, F.~Ju, B.~Chen, H.~Guo, D.~Wang, Y.~Wang, and H.~Wu, ``Analysis of a
  novel manipulator with low melting point alloy initiated stiffness variation
  and shape detection for minimally invasive surgery,'' \emph{Industrial Robot:
  the international journal of robotics research and application}, vol.~48,
  no.~2, pp. 247--258, 2021.

\bibitem{saavedra2013variable}
E.~I. Saavedra~Flores, M.~I. Friswell, and Y.~Xia, ``Variable stiffness
  biological and bio-inspired materials,'' \emph{Journal of Intelligent
  Material Systems and Structures}, vol.~24, no.~5, pp. 529--540, 2013.

\bibitem{zhou2022bioinspired}
J.~Zhou, H.~Cao, W.~Chen, S.~S. Cheng, and Y.-H. Liu, ``Bioinspired soft wrist
  based on multicable jamming with hybrid motion and stiffness control for
  dexterous manipulation,'' \emph{IEEE/ASME Transactions on Mechatronics},
  vol.~28, no.~3, pp. 1256--1267, 2022.

\bibitem{zhou2020adaptive}
J.~Zhou, Y.~Chen, Y.~Hu, Z.~Wang, Y.~Li, G.~Gu, and Y.~Liu, ``Adaptive variable
  stiffness particle phalange for robust and durable robotic grasping,''
  \emph{Soft robotics}, vol.~7, no.~6, pp. 743--757, 2020.

\bibitem{arleo2023variable}
L.~Arleo, L.~Lorenzon, and M.~Cianchetti, ``Variable stiffness linear actuator
  based on differential drive fiber jamming,'' \emph{IEEE Transactions on
  Robotics}, vol.~39, no.~6, pp. 4429--4442, 2023.

\bibitem{zeng2020parallel}
X.~Zeng, C.~Hurd, H.-J. Su, S.~Song, and J.~Wang, ``A parallel-guided compliant
  mechanism with variable stiffness based on layer jamming,'' \emph{Mechanism
  and Machine Theory}, vol. 148, p. 103791, 2020.

\bibitem{clark2019assessing}
A.~B. Clark and N.~Rojas, ``Assessing the performance of variable stiffness
  continuum structures of large diameter,'' \emph{IEEE Robotics and Automation
  Letters}, vol.~4, no.~3, pp. 2455--2462, 2019.

\bibitem{aktacs2021modeling}
B.~Akta{\c{s}}, Y.~S. Narang, N.~Vasios, K.~Bertoldi, and R.~D. Howe, ``A
  modeling framework for jamming structures,'' \emph{Advanced Functional
  Materials}, vol.~31, no.~16, p. 2007554, 2021.

\bibitem{li2019flexible}
C.~Li, X.~Gu, X.~Xiao, C.~M. Lim, and H.~Ren, ``Flexible robot with variable
  stiffness in transoral surgery,'' \emph{IEEE/ASME Transactions on
  Mechatronics}, vol.~25, no.~1, pp. 1--10, 2019.

\bibitem{shen2023design}
D.~Shen, Q.~Zhang, Y.~Han, C.~Tu, and X.~Wang, ``Design and development of a
  continuum robot with switching-stiffness,'' \emph{Soft Robotics}, vol.~10,
  no.~5, pp. 1015--1027, 2023.

\bibitem{yang2020geometric}
C.~Yang, S.~Geng, I.~Walker, D.~T. Branson, J.~Liu, J.~S. Dai, and R.~Kang,
  ``Geometric constraint-based modeling and analysis of a novel continuum robot
  with shape memory alloy initiated variable stiffness,'' \emph{The
  International Journal of Robotics Research}, vol.~39, no.~14, pp. 1620--1634,
  2020.

\bibitem{lynch2017modern}
K.~M. Lynch and F.~C. Park, \emph{Modern Robotics}.\hskip 1em plus 0.5em minus
  0.4em\relax Cambridge University Press, 2017.

\bibitem{abbasifard2014survey}
M.~R. Abbasifard, B.~Ghahremani, and H.~Naderi, ``A survey on nearest neighbor
  search methods,'' \emph{International Journal of Computer Applications},
  vol.~95, no.~25, 2014.

\bibitem{bentley1975multidimensional}
J.~L. Bentley, ``Multidimensional binary search trees used for associative
  searching,'' \emph{Communications of the ACM}, vol.~18, no.~9, pp. 509--517,
  1975.

\bibitem{cormen2022introduction}
T.~H. Cormen, C.~E. Leiserson, R.~L. Rivest, and C.~Stein, \emph{Introduction
  to algorithms}.\hskip 1em plus 0.5em minus 0.4em\relax MIT press, 2022.

\bibitem{beer1992mechanics}
F.~P. Beer, E.~R. Johnston, J.~T. DeWolf, D.~F. Mazurek, and S.~Sanghi,
  \emph{Mechanics of materials}.\hskip 1em plus 0.5em minus 0.4em\relax
  mcgraw-Hill New York, 1992, vol.~1.

\bibitem{chen2021tele}
W.~Chen, J.~Zhou, S.~S. Cheng, Y.~Lu, F.~Zhong, Y.~Gao, Y.~Wang, L.~Xue, M.~C.
  Tong, and Y.-H. Liu, ``Tele-operated oropharyngeal swab (toos) robot enabled
  by tss soft hand for safe and effective sampling,'' \emph{IEEE Transactions
  on Medical Robotics and Bionics}, vol.~3, no.~4, pp. 1040--1053, 2021.

\bibitem{zhou2021bio}
J.~Zhou, W.~Chen, S.~S. Cheng, L.~Xue, M.~C. Tong, and Y.~Liu, ``Bio-inspired
  soft (bis) hand for tele-operated covid-19 oropharyngeal (op) swab
  sampling,'' in \emph{2021 IEEE International Conference on Robotics and
  Biomimetics (ROBIO)}.\hskip 1em plus 0.5em minus 0.4em\relax IEEE, 2021, pp.
  80--86.

\bibitem{farooq2018stiffness}
M.~U. Farooq and S.~Y. Ko, ``A stiffness-changing continuum robotic manipulator
  for possible use in mri-guided neurosurgical interventions,'' in \emph{IEEE
  International Conference on Biomedical Robotics and Biomechatronics
  (Biorob)}, 2018, pp. 1260--1265.

\bibitem{brancadoro2019toward}
M.~Brancadoro, M.~Manti, F.~Grani, S.~Tognarelli, A.~Menciassi, and
  M.~Cianchetti, ``Toward a variable stiffness surgical manipulator based on
  fiber jamming transition,'' \emph{Frontiers in Robotics and AI}, vol.~6,
  p.~12, 2019.

\bibitem{jiang2012design}
A.~Jiang, G.~Xynogalas, P.~Dasgupta, K.~Althoefer, and T.~Nanayakkara, ``Design
  of a variable stiffness flexible manipulator with composite granular jamming
  and membrane coupling,'' in \emph{IEEE/RSJ International Conference on
  Intelligent Robots and Systems}, 2012, pp. 2922--2927.

\bibitem{li2020single}
C.~Li, X.~Xiao, X.~Gu, F.~Jie, S.~Sangeetha, T.~Z.~W. Joelle, T.~T.~W. Kiat,
  and H.~Ren, ``Single-port multichannel multi-degree-of-freedom robot with
  variable stiffness for natural orifice transluminal endoscopic surgery,''
  \emph{Flexible robotics in medicine: a design journey of motion generation
  mechanisms and biorobotic system development}, vol. 389, 2020.

\bibitem{liu2021positive}
T.~Liu, H.~Xia, D.-Y. Lee, A.~Firouzeh, Y.-L. Park, and K.-J. Cho, ``A positive
  pressure jamming based variable stiffness structure and its application on
  wearable robots,'' \emph{IEEE Robotics and Automation Letters}, vol.~6,
  no.~4, pp. 8078--8085, 2021.

\bibitem{sadati2015stiffness}
S.~H. Sadati, Y.~Noh, S.~Elnaz~Naghibi, K.~Althoefer, and T.~Nanayakkara,
  ``Stiffness control of soft robotic manipulator for minimally invasive
  surgery (mis) using scale jamming,'' in \emph{Intelligent Robotics and
  Applications: 9th International Conference, ICIRA 2015, Portsmouth, UK,
  August 24--27, 2015, Proceedings, Part III}.\hskip 1em plus 0.5em minus
  0.4em\relax Springer, 2015, pp. 141--151.

\bibitem{sadati2018three}
S.~H. Sadati, L.~Sullivan, I.~D. Walker, K.~Althoefer, and T.~Nanayakkara,
  ``Three-dimensional-printable thermoactive helical interface with
  decentralized morphological stiffness control for continuum manipulators,''
  \emph{IEEE Robotics and Automation Letters}, vol.~3, no.~3, pp. 2283--2290,
  2018.

\bibitem{sun2020hybrid}
C.~Sun, L.~Chen, J.~Liu, J.~S. Dai, and R.~Kang, ``A hybrid continuum robot
  based on pneumatic muscles with embedded elastic rods,'' \emph{Proceedings of
  the Institution of Mechanical Engineers, Part C: Journal of Mechanical
  Engineering Science}, vol. 234, no.~1, pp. 318--328, 2020.

\bibitem{moses2013continuum}
M.~S. Moses, M.~D. Kutzer, H.~Ma, and M.~Armand, ``A continuum manipulator made
  of interlocking fibers,'' in \emph{IEEE International Conference on Robotics
  and Automation}, 2013, pp. 4008--4015.

\bibitem{zhao2019soft}
Y.~Zhao, Y.~Shan, J.~Zhang, K.~Guo, L.~Qi, L.~Han, and H.~Yu, ``A soft
  continuum robot, with a large variable-stiffness range, based on jamming,''
  \emph{Bioinspiration \& biomimetics}, vol.~14, no.~6, p. 066007, 2019.

\bibitem{zhu2021controllable}
X.~Zhu and H.~Hu, ``A controllable stiffness robotics for natural orifice
  transluminal endoscopic surgery,'' in \emph{World Conference on Mechanical
  Engineering and Intelligent Manufacturing (WCMEIM)}.\hskip 1em plus 0.5em
  minus 0.4em\relax IEEE, 2021, pp. 45--48.

\end{thebibliography}
%



%
\begin{IEEEbiography}
[{\includegraphics[width=1in,height=1.25in]{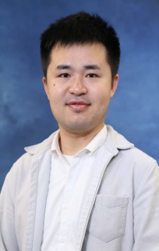}}]
{Jianshu Zhou} (Member, IEEE) received his Ph.D. in Mechanical Engineering from The University of Hong Kong in 2020. He is currently a postdoctoral researcher at the Mechanical Systems Control Lab, Department of Mechanical Engineering, University of California, Berkeley. Prior to this, he served as a research assistant professor at The Chinese University of Hong Kong. His research interests include robotics, dexterous hands, adaptive stiffness robots, human-robot interface, and embodied intelligence.
\end{IEEEbiography}

\begin{IEEEbiography}
[{\includegraphics[width=1in,height=1.25in]{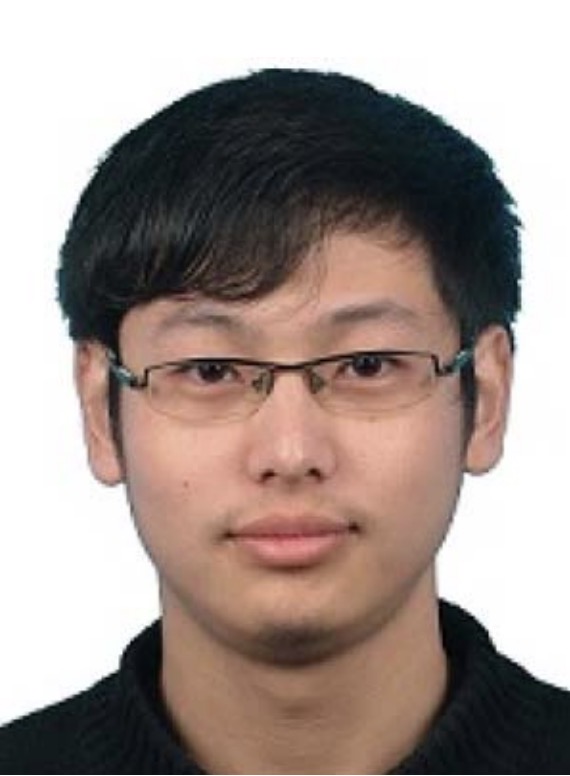}}]
{Wei Chen}  received the B.E. degree in Computer Science and Technology from Zhengzhou University, Zhengzhou, China, in 2012, and the M.S. degree in Mechanical and Automation Engineering from The Chinese University of Hong Kong, Hong Kong, in 2021. He is currently pursuing the Ph.D. degree in the Department of Mechanical and Automation Engineering at The Chinese University of Hong Kong, Hong Kong SAR. His research interests include medical robotics and soft robotics.
\end{IEEEbiography}

\begin{IEEEbiography}
[{\includegraphics[width=1in,height=1.25 in]{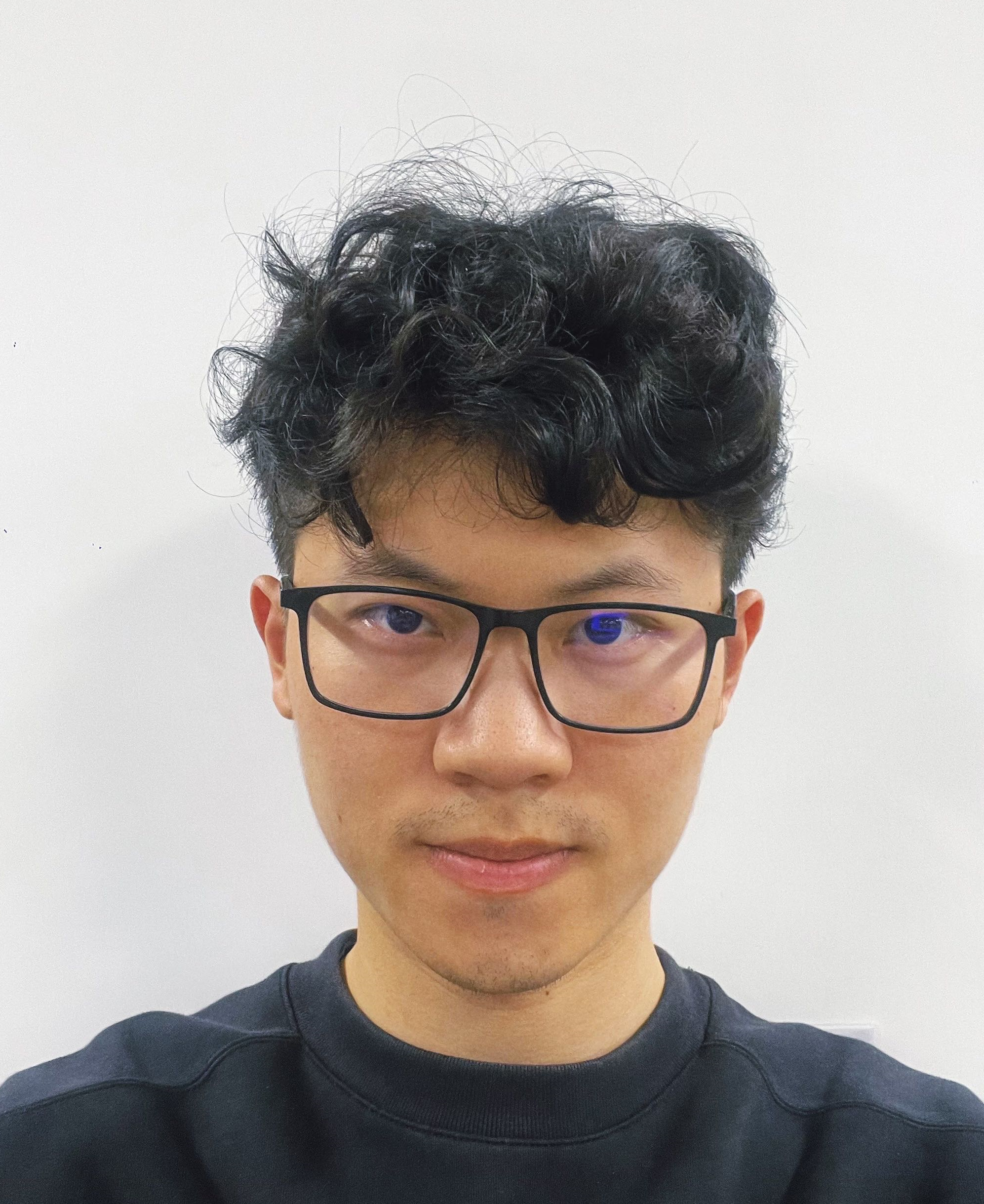}}]
{Junda Huang} received the B.Eng. degrees from the University of Science and Technology of China (USTC) in 2020. Currently, he is pursing the Ph.D. degree at the CUHK. His research interest includes dexterous hand, robotic grasping and manipulation, teleoperation, and imitation learning.
\end{IEEEbiography}

\begin{IEEEbiography}
[{\includegraphics[width=1in,height=1.25in]{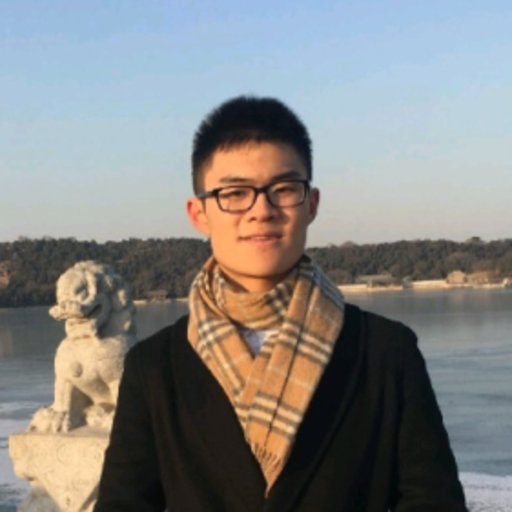}}]
{Boyuan Liang} received the B.Sc. (Hons) degree in Applied Mathematics from National University of Singapore in 2021. He is currently pursuing Ph.D. degree in Mechanical Engineering at University of California, Berkeley. His research interest lies in robotics, mechatronics systems, teleoperation and contact modeling.
\end{IEEEbiography}

\begin{IEEEbiography}
[{\includegraphics[width=1in,height=1.25in]{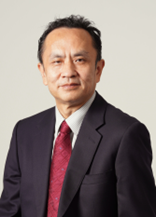}}]
{Yunhui Liu} (Fellow, IEEE) received his Ph.D. in Applied Mathematics and Information Physics from the University of Tokyo. After working as a Research Scientist at the Electrotechnical Laboratory of Japan, he joined The Chinese University of Hong Kong (CUHK) in 1995. He is currently the Choh-Ming Li Professor of Mechanical and Automation Engineering and the Director of the T Stone Robotics Institute.
Dr. Liu also serves as the Director/CEO of the Hong Kong Centre for Logistics Robotics, sponsored by the InnoHK programme of the HKSAR government. He is an adjunct professor at the State Key Laboratory of Robotics Technology and System at Harbin Institute of Technology, China.
With over 500 papers published in refereed journals and conference proceedings, Dr. Liu was listed as a Highly Cited Author (Engineering) by Thomson Reuters in 2013. His research interests span visual servoing, logistics robotics, medical robotics, multi-fingered grasping, mobile robots, and machine intelligence.
Dr. Liu has received numerous research awards from prestigious international journals, conferences in robotics and automation, and government agencies. He was the Editor-in-Chief of Robotics and Biomimetics, an Associate Editor of the IEEE Transactions on Robotics and Automation, and the General Chair of the 2006 IEEE/RSJ International Conference on Intelligent Robots and Systems. He is also an IEEE Fellow.
\end{IEEEbiography}

\begin{IEEEbiography}
[{\includegraphics[width=1in,height=1.25in]{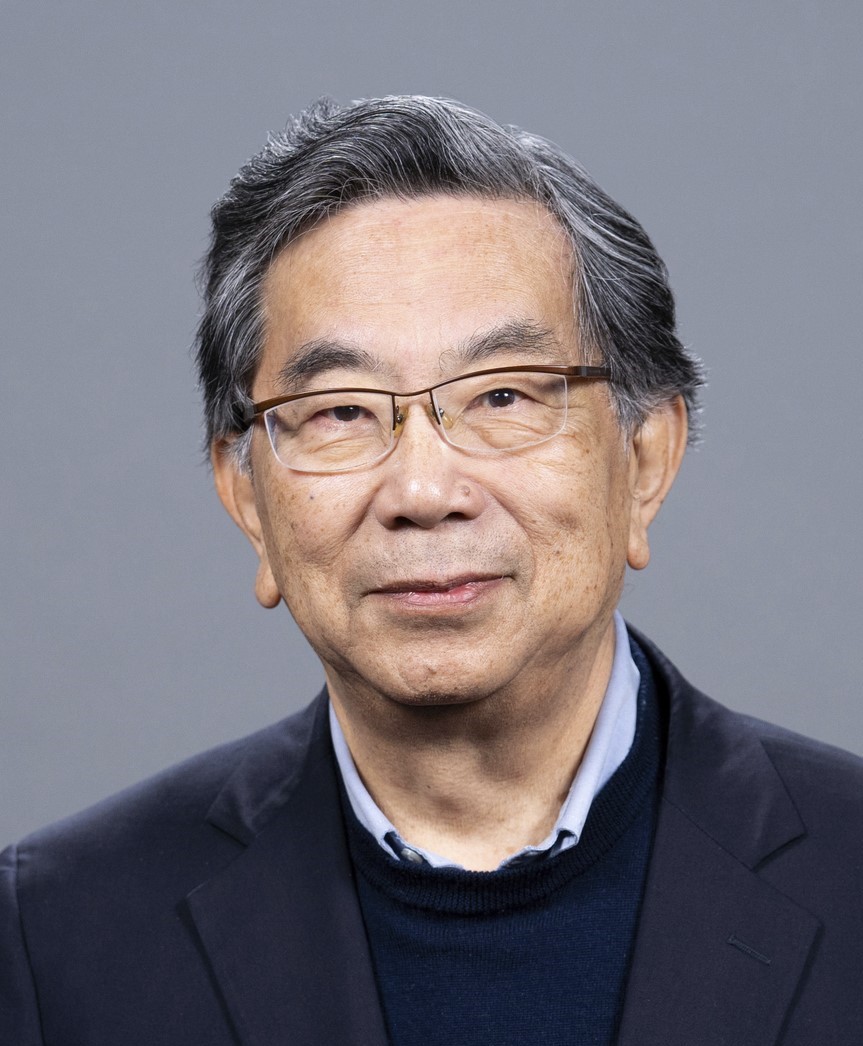}}]
{Masayoshi Tomizuka} (Life Fellow, IEEE/ASME) was born in Tokyo, Japan, in 1946. He received the B.S. and M.S. degrees in mechanical engineering from Keio University, Tokyo, Japan, in 1968 and 1970, respectively, and the Ph.D. degree in mechanical engineering from the Massachusetts Institute of Technology, Cambridge, MA, USA, in February 1974.,In 1974, he joined the Faculty of the Department of Mechanical Engineering, University of California at Berkeley, Berkeley, CA, USA, where he is currently the Cheryl and John Neerhout, Jr., Distinguished Professorship Chair. At UC Berkeley, he teaches courses in dynamic systems and controls. From 2002 to 2004, he was the Program Director of the Dynamic Systems and Control Program of the Civil and Mechanical Systems Division of NSF. His current research interests are optimal and adaptive control, digital control, signal processing, motion control, and control problems related to robotics, machining, manufacturing, information storage devices, and vehicles.,Dr. Tomizuka was the Technical Editor of the ASME Journal of Dynamic Systems, Measurement and Control, J-DSMC (1988–1993), Editor-in-Chief of the IEEE/ASME Transactions on Mechatronics (1997–1999), and Associate Editor for the Journal of the International Federation of Automatic Control, and Automatica. He was the General Chairman of the 1995 American Control Conference, and was the President of the American Automatic Control Council (1998–1999). He is a Life Fellow of the ASME and a Fellow of the International Federation of Automatic Control (IFAC) and the Society of Manufacturing Engineers. He was the recipient of the Best J-DSMC Best Paper Award (1995, 2010), DSCD Outstanding Investigator Award (1996), Charles Russ Richards Memorial Award (ASME, 1997), Rufus Oldenburger Medal (ASME, 2002), John R. Ragazzini Award (AACC, 2006), Richard E. Bellman Control Heritage Award (AACC, 2018), Honda Medal (ASME, 2019), and Nathaniel B. Nichols Medal (IFAC, 2020). He is a member of the National Academy of Engineering.

\end{IEEEbiography}









\end{document}